\documentclass{article}
\usepackage[T1]{fontenc}
\usepackage{amsmath}
\usepackage{amsthm,amssymb,amsfonts,bbm,bm}
\usepackage{natbib}
\usepackage{fullpage}
\usepackage{mathtools}
\usepackage{hyperref}
\usepackage{url}
\usepackage{xcolor}
\usepackage{subfigure}
\usepackage{multirow}

\newtheorem{theorem}{Theorem}[section]
\newtheorem{lemma}{Lemma}[section]

\numberwithin{equation}{section}

\newcommand{\bmu}{\boldsymbol{\mu}}
\newcommand{\bmus}{\boldsymbol{\mu_*}}
\newcommand{\bu}{\boldsymbol{U}}
\newcommand{\buq}{\boldsymbol{U_q}}
\newcommand{\bus}{\boldsymbol{U_*}}
\newcommand{\bsigma}{\boldsymbol{\Sigma}}
\newcommand{\bsigmas}{\boldsymbol{\Sigma_*}}
\newcommand{\bsigmasi}{\boldsymbol{\Sigma_*}^{-1}}
\newcommand{\bsigmaj}{\boldsymbol{\Sigma_j}}
\newcommand{\bw}{\boldsymbol{W}}
\newcommand{\blambda}{\boldsymbol{\Lambda}}
\newcommand{\blambdas}{\boldsymbol{\Lambda_*}}
\newcommand{\blambdaj}{\boldsymbol{\Lambda^{(j)}}}
\newcommand{\blambdaq}{\boldsymbol{\Lambda_q}}
\newcommand{\blambdam}{\boldsymbol{\Lambda_M}}

\newcommand{\bp}{\boldsymbol{P}^{-1}}

\hypersetup{
  colorlinks,
  citecolor=blue,
  linkcolor=red,
  urlcolor=blue}
  
\title{Understanding Adversarial Robustness Against On-manifold Adversarial Examples}
\author{%
  Jiancong Xiao$^1$, Liusha Yang$^3$, Yanbo Fan$^{2,}\thanks{Correspondence Authors.}$\ , Jue Wang$^{2}$, Zhi-Quan Luo$^{1,3,*}$\\
   $^1$The Chinese University of Hong Kong, Shenzhen; \\ 
   $^2$Tencent AI Lab; $^3$Shenzhen Research Institute of Big Data\\
  \texttt{jiancongxiao@link.cuhk.edu.cn, yangliusha@sribd.cn,} \\
  \texttt{fanyanbo0124@gmail.com, arphid@gmail.com, luozq@cuhk.edu.cn}
}
\date{}
\begin{document}

\maketitle

\begin{abstract}
  Deep neural networks (DNNs) are shown to be vulnerable to adversarial examples. A well-trained model can be easily attacked by adding small perturbations to the original data. One of the hypotheses of the existence of the adversarial examples is the off-manifold assumption: adversarial examples lie off the data manifold. However, recent research showed that on-manifold adversarial examples also exist. In this paper, we revisit the off-manifold assumption and want to study a question: at what level is the poor performance of neural networks against adversarial attacks due to on-manifold adversarial examples? Since the true data manifold is unknown in practice, we consider two approximated on-manifold adversarial examples 
 on both real and synthesis datasets. On real datasets, we show that on-manifold adversarial examples have greater attack rates than off-manifold adversarial examples on both standard-trained and adversarially-trained models. On synthetic datasets, theoretically, We prove that on-manifold adversarial examples are powerful, yet adversarial training focuses on off-manifold directions and ignores the on-manifold adversarial examples. Furthermore, we provide analysis to show that the properties derived theoretically can also be observed in practice. Our analysis suggests that on-manifold adversarial examples are important, and we should pay more attention to on-manifold adversarial examples for training robust models.
\end{abstract}

\section{Introduction}

In recent years, deep neural networks (DNNs) \citep{krizhevsky2012imagenet,hochreiter1997long} have become popular and successful in many machine learning tasks. They have been used in different problems with great success. But DNNs are shown to be vulnerable to adversarial examples \citep{szegedy2013intriguing,goodfellow2014explaining}. A well-trained model can be easily attacked by adding small perturbations to the images. One of the hypotheses of the existence of adversarial examples is the off-manifold assumption \citep{szegedy2013intriguing}: 

\begin{center}
\emph{Clean data lies in a low-dimensional manifold. Even though the adversarial examples are close to the clean data, they lie off the underlying data manifold.}
\end{center}

DNNs only fit the data in the manifold and perform badly on adversarial examples out of the manifold. Many research support this point of view. Pixeldefends \citep{song2017pixeldefend} leveraged a generative model to show that adversarial examples lie in a low probability region of the data distribution. The work of \citep{gilmer2018adversarial,khoury2018geometry} studied the geometry of adversarial examples. It shows that adversarial examples are related to the high dimension of the data manifold and they are conducted in the directions off the data manifold. The work of \citep{ma2018characterizing} useed Local Intrinsic Dimensionality (LID) to characterize the adversarial region and argues that the adversarial subspaces are of low probability, and lie off (but are close to) the data submanifold.

One of the most effective approaches to improve the adversarial robustness of DNNs is to augment adversarial examples to the training set, i.e., adversarial training. However, the performance of adversarially-trained models are still far from satisfactory. Based on the off-manifold assumption, previous studies provided an explanation on the poor performance of adversarial training that the adversarial data lies in a higher-dimensional manifold. DNNs can work well on a low dimensional manifold but not on a high dimensional manifold. This is discussed in the paper we mentioned above \citep{gilmer2018adversarial,khoury2018geometry}.
There are also other interpretations of the existence of adversarial examples, see Sec. \ref{sec:related_work}.

In recent years, researchers found that on-manifold adversarial examples also exist. They can fool the target models \citep{lin2020dual}, boost clean generalization \citep{stutz2019disentangling}, improve uncertainty calibration \citep{patel2021manifold}, and improve model compression \citep{kwon2021improving}. Therefore, the off-manifold assumption may not be a perfect hypothesis explaining the existence of adversarial examples. It motivates us to revisit the off-manifold assumption. Specifically, we study the following question:
\begin{center}
\emph{At what level is the poor performance of neural networks against adversarial attacks due to on-manifold adversarial examples?}
\end{center} 

The main difficulty in studying this question is that the true data manifold is unknown in practice. It is hard to obtain the true on-manifold adversarial examples. To have a closer look at on-manifold adversarial examples, we consider two approximated on-manifold adversarial examples, and we consider both real and synthetic datasets.

\paragraph{Approximate on-manifold adversarial examples} We use two approaches to approximate the on-manifold adversarial examples. The first one is generative adversarial examples (Gen-AE). Generative models, such as generative adversarial networks (GAN) \citep{goodfellow2014generative} and variational autoencoder (VAE) \citep{kingma2013auto}, are used to craft adversarial examples. Since the generative model is an approximation of the data manifold, the crafted data lies in the data manifold. We perform perturbation in the latent space of the generative model. Since the first approach relies on the quality of the generative models, we consider the second approach: using the eigenspace of the training dataset to approximate the data manifold, which we call eigenspace adversarial examples (Eigen-AE). In this method, the crafted adversarial examples are closer to the original samples. For more details, see Section \ref{onmanifold}.

\begin{table}[]
    \label{tdata}
    \centering
    \caption{Summary of the main results of our paper. In Sec. \ref{onmanifold}, we show that (approximated) on-manifold adversarial examples (Gen-AE, Eigen-AE) have higher attack rates than off-manifold adversarial examples. In Sec. \ref{sec:ta}, we provide a theoretical analysis of on-manifold attacks on GMMs, where the true data manifold is known. In Sec. \ref{sec:exper}, we provide further analysis to show the similarity of these four cases.}
    \resizebox{\linewidth}{!}{%
    \begin{tabular}{ccc}
    \hline
         &  Real datasets & Synthetic datasets (GMMs, known manifold)\\
    \hline
      \multirow{2}{*}{Gen-AE}& \multirow{2}{*}{higher attack rates (Table \ref{t2})}  & Excess risk: Thm. \ref{thm:er2} \\
        & & Adversarial distribution shift: Thm. \ref{thm:alp2}\\
        \hline
      \multirow{2}{*}{Eigen-AE} & \multirow{2}{*}{higher attack rates  (Fig. \ref{fig1})} & Excess risk: Thm. \ref{thm:er3}\\
     &  & Adversarial distribution shift: Thm: \ref{thm:robust}\\
    \hline
    \end{tabular}}
\end{table}
To start, we provide experiments of on-manifold attacks on both standard-trained and adversarially-trained models on MNIST, CIFAR-10, CIFAR-100, and ImageNet. The experiments show that on-manifold attacks are powerful, with higher attack rates than off-manifold adversarial examples. It help justifies that the off-manifold assumption might not be a perfect hypothesis of the existence of adversarial examples.

The experiments motivate us to study on-manifold adversarial examples from a theoretical perspective. Since the true data manifold is unknown in practice, we provide a theoretical study on synthetic datasets using Gaussian mixture models (GMMs). In this case, the true data manifold is given. We study the \emph{excess risk} (Thm. \ref{thm:er2} and Thm. \ref{thm:er3}) and \emph{adversarial distribution shift} (Thm. \ref{thm:alp2} and Thm. \ref{thm:robust}) of these two types of on-manifold adversarial examples. Our main technical contribution is providing closed-form solutions to the min-max problems of on-manifold adversarial training. Our theoretical results show that on-manifold adversarial examples incur a large excess risk and foul the target models. Compared to regular adversarial attacks, we show how adversarial training focuses on off-manifold directions and ignores the on-manifold adversarial examples. 

Finally, we provide a comprehensive analysis to connect the four settings (two approximate on-manifold attacks on both real and synthetic datasets). We show the similarity of these four cases: 1) the attacks directions of Gen-AE and Eigen-AE are similar on common datasets, and 2) the theoretical properties derived using GMMs (Thm. \ref{thm:er2} to Thm. \ref{thm:robust}) can also be observed in practice.

Based on our study, we find that on-manifold adversarial examples are important for adversarial robustness. We emphasize that we should pay more attention to on-manifold adversarial examples for training robust models. Our contributions are listed as follows:

\begin{itemize}
\item We develop two approximate on-manifold adversarial attacks to take a closer look at on-manifold adversarial examples. 
	\item We provide comprehensive analyses of on-manifold adversarial examples empirically and theoretically. We summarize our main results in Table \ref{tdata}. Our results suggest the importance of on-manifold adversarial examples, and we should pay more attention to on-manifold adversarial examples to train robust models.
	\item Technical contributions: our main technical contribution is providing the closed-form solutions to the min-max problems of on-manifold adversarial training (Thm. \ref{thm:alp2} and \ref{thm:robust}) in GMMs. We also provide the upper and lower bounds of excess risk (Thm. \ref{thm:er2} and \ref{thm:er3}) of on-manifold attacks.
\end{itemize}

\section{Related Work}

\label{sec:related_work}
\paragraph{Attack} Adversarial examples for deep neural networks were first intruduced in \citep{szegedy2013intriguing}. However, adversarial machine learning or robust machine learning has been studied for a long time \citep{biggio2018wild}. In the setting of white box attack \citep{kurakin2016adversarial,papernot2016limitations,moosavi2016deepfool,carlini2017towards}, the attackers have fully access to the model (weights, gradients, etc.). In black box attack \citep{chen2017zoo,su2019one,ilyas2018prior}, the attackers have limited access to the model. First order optimization methods, which use the gradient information to craft adversarial examples, such as PGD \citep{madry2017towards}, are widely used for white box attack. Zeroth-order optimization methods \citep{chen2017zoo} are used in black box setting. \citep{li2019nattack} improved the query efficiency in black-box attack. \emph{Generative adversarial examples:} Generative models have been used to craft adversarial examples \citep{xiao2018generating,song2018constructing,kos2018adversarial}. The adversarial examples are more natural \citep{zhao2017generating}.

\paragraph{Defense}
Training algorithms against adversarial attacks can be subdivided into the following categories. \emph{Adversarial training:} The training data is augmented with adversarial examples to make the models more robust \citep{madry2017towards,szegedy2013intriguing,tramer2017ensemble}. \emph{Preprocessing:} Inputs or hidden layers are quantized, projected onto different sets or other preprocessing methods \citep{buckman2018thermometer,guo2017countering,kabilan2018vectordefense}. \emph{Stochasticity:} Inputs or hidden activations are randomized \citep{prakash2018deflecting,dhillon2018stochastic,xie2017mitigating}. However, some of them are shown to be useless defenses given by obfuscated gradients \citep{athalye2018obfuscated}. Adaptive attack \citep{tramer2020adaptive} is used for evaluating defenses to adversarial examples. \citep{xiao2022adversarial} consider the Rademacher complexity in adversarial training case.

\section{On-manifold Adversarial Attacks}
\label{sec:ls}
\paragraph{Adversarial Attacks} Given a classifier $f_\theta$ and a sample data $(x,y)$. The goal of \emph{regular adversarial attacks} is to find an adversarial example $x'$ to foul the classifier $f_\theta$. A widely studied norm-based adversarial attack is to solve the optimization problem $\max_{\|x-x'\|\leq\varepsilon}\ell(f_\theta(x'),y),$ where $\ell(\cdot,\cdot)$ is the loss function and $\varepsilon$ is the perturbation intensity. 
In this paper, we focus on on-manifold adversarial attacks. We must introduce additional constraints to restrict $x'$ in the data manifold and preserve the label $y$. In practice, the true data manifold is unknown. Assuming that the true data manifold is a push-forward function $p(z):\mathcal{Z}\rightarrow \mathcal{X}$, where $\mathcal{Z}$ is a low dimensional Euclidean space and $\mathcal{X}$ is the support of the data distribution. One way to approximate $p(z)$ is to use a generative model $G(z)$ such that $G(z)\approx p(z)$. The second way is to consider the Taylor expansion of $p(z)$ and use the first-order term to approximate $p(z)$. They correspond to the following two approximated on-manifold adversarial examples. 

\paragraph{Generative Adversarial Examples} One method to approximate the data manifold is to use generative models, such as GAN and VAE. Let $G:\mathcal{Z}\rightarrow\mathcal{X}$ be a generative model and $I:\mathcal{X}\rightarrow\mathcal{Z}$ be the inverse mapping of $G(z)$. Generative adversarial attack can be formulated as the following problem
\begin{equation}
\label{eq:lsat}
\max_{\|z'-I(x)\|\leq\varepsilon}\ell(f_\theta(G(z')),y).
\end{equation}
Then, $x'=G(z')$ is an (approximate) on-manifold adversarial example. In this method, $\|x-x'\|$ can be large. To preserve the label $y$, we use the conditional generative models (e.g. C-GAN \citep{mirza2014conditional} and C-VAE \citep{sohn2015learning}, i.e. the generator $G_y(z)$ and inverse mapping $I_y(x)$ are conditioned on the label $y$. In the experiments, we use two widely used gradient-based attack algorithms, \emph{fast gradient sign method} (FGSM) \citep{goodfellow2014explaining} and \emph{projected gradient descend} (PGD) \citep{madry2017towards} in the latent space $\mathcal{Z}$, which we call GFGSM and GPGD, respectively.

\paragraph{Eigenspace Adversarial Examples} The on-manifold adversarial examples found by the above method are not norm-bounded. To study norm-based on-manifold adversarial examples, we use the eigenspace to approximate the data manifold. Consider the following problem
\begin{equation}
\label{eq:eae}
\begin{aligned}
\max_{x'}&\quad \ell(f_\theta(x'),y)\\
s.t.&\quad A_y^\perp(x-x')=0,\  \|x-x'\|\leq\varepsilon
\end{aligned}
\end{equation}
where the rows of $A_y$ is the eigenvectors corresponds to the top eigenvalues of the co-variance matrix of the training data with label $y$. Then, $x'$ is restricted in the eigenspace. A baseline algorithm for standard adversarial attack is PGD attack \citep{madry2017towards}. To make a fair comparison, we should use the same algorithm to solve the attack problem. Notice that the inner maximization problem in Eq. (\ref{eq:eae}) cannot be solved by PGD in general. Because the projection step is an optimization problem. We consider $\ell_2$ adversarial attacks in this case. Then, Eq. (\ref{eq:eae}) is equivalent to
\begin{equation}
\label{eq:l2}
\begin{aligned}
\max_{\|\Delta z\|_2\leq\varepsilon,}&\quad \ell(f(x+\Delta z^T A_y),y).
\end{aligned}
\end{equation}
We can use PGD to find eigenspace adversarial examples. In this case, the perturbation constraint is a subset of the constraint of regular adversarial attacks. For comparison, we consider the case that the rows of $A_y$ is the eigenvectors corresponds to the bottom eigenvalues. We call 
this type of on-manifold and off-manifold adversarial examples as top eigenvectors and bottom eigenvectors subspace adversarial examples.

\paragraph{Target Models} In this paper, we aim to study the performance of on-manifold adversarial attacks on both standard-trained model and adversarially-trained model. For standard training, only the original data is used in training. For adversarial training, adversarial examples are augmented to the training dataset. For ablation studies, we also consider on-manifold adversarial training, \emph{i.e.,} on-manifold adversarial examples crafted by the above mentioned algorithms are augmented to the training dataset.

\section{Warm-up: Performance of On-manifold Adversarial Attacks}
\label{sec:perform}
In this section, we provide experiments in different settings to study the performance of on-manifold adversarial attacks.
\label{onmanifold}
\begin{table*}[htbp]
	\centering
	\caption{Test accuracy of different defense algorithms (PGD-AT, GPGD-AT, and joint-PGD-AT) against different attacks (regular attacks (FGSM, PGD) and generative attacks (GFGSM, GPGD) on MNIST, CIFAR-10, and ImageNet.}
	\resizebox{\linewidth}{!}{%
	\begin{tabular}{cccccc}
		\hline
		MNIST & clean data & FGSM-Attack & PGD-Attack& GFGSM-Attack& GPGD-Attack\\
		\hline
		Std training &  \textbf{98.82\%}&47.38\%&3.92\%&42.37\%& 10.74\%\\
		PGD-AT & 98.73\%&96.50\%&95.51\%&52.17\%&15.53\%\\
		GPGD-AT&98.63\%&20.63\%&2.11\%&\textbf{99.66\%}&\textbf{96.78\%}\\
		joint-PGD-AT & 98.45\%&\textbf{97.31\%}&\textbf{95.70\%}&\underline{99.27\%}&\underline{96.03\%}\\
		\hline
		CIFAR-10 & clean data & FGSM-Attack & PGD-Attack& GFGSM-Attack& GPGD-Attack\\
		\hline
		Std training & \textbf{91.80\%}&15.08\%&5.39\%&7.07\%& 3.41\%\\
		PGD-AT & 80.72\%&\textbf{56.42\%}&\textbf{50.18\%}&14.27\%&8.51\%\\
		GPGD-AT&78.93\%&10.64\%&3.21\%&40.18\%&26.66\%\\
		joint-PGD-AT & 79.21\%&\underline{50.19\%}&\underline{49.77\%}&\textbf{42.87\%}&\textbf{28.54\%}\\
		\hline
		ImageNet & clean data & FGSM-Attack & PGD-Attack& GFGSM-Attack& GPGD-Attack\\
		\hline
		Std training &  74.72\%&2.59\%&0.00\%&/& 0.26\%\\
		PGD-AT & 73.31\%&48.02\%&\textbf{38.88\%}&/&7.23\%\\
		GPGD-AT&78.10\%&21.68\%&0.03\%&/&\textbf{27.53\%}\\
		joint-PGD-AT & \textbf{77.96\%}&\textbf{49.12\%}&\underline{37.86\%}&/&\underline{20.53\%}\\
		\hline
	\end{tabular}
	}
	\label{t2}
\end{table*}

\subsection{Experiments of Generative Adversarial Attacks}

To study the performance of generative adversarial attacks, we compare the performance of different attacks (FGSM, PGD, GFGSM, and GPGD) versus different models (standard training, PGD-adversarial training, and GPGD-adversarial training). 

\paragraph{Experiments setup} In this section we report our experimental results on training LeNet on MNIST \citep{lecun1998gradient}, ResNet18 \citep{he2016deep} on CIFAR-10 \citep{krizhevsky2009learning}, and ResNet50 on ImageNet \citep{deng2009imagenet}. The experiments on CIFAR-100 are discussed in Appendix B.3. On MNIST, we use $\varepsilon=0.3$ and 40 steps PGD for adversarial training and use  $\varepsilon=1$ and 40 steps PGD for generative adversarial training. On CIFAR-10 and CIFAR-100, we use  $\varepsilon=8/255$, PGD-10 for adversarial training, and $\varepsilon=0.1$, PGD-10 for generative adversarial training. On ImageNet, we adopt the settings in \citep{lin2020dual} and use  $\varepsilon=4/255$, PGD-5 for adversarial training, and $\varepsilon=0.02$, PGD-5 for generative adversarial training. The choice of $\epsilon$ in the latent space is based on the quality of the generative examples. In the outer minimization, we use the SGD optimizer with momentum 0.9 and weight decay $5 \cdot 10^{-4}$. For reference Details of the hyperparameters setting are in Appendix B.1.

\paragraph{Generative AT cannot defend regular attack, and vice versa} In Table \ref{t2}, the test accuracy of GPGD-AT vs PGD-attack is 3.21\%, which means that on-manifold adversarial training cannot defend a regular norm-based attack. Similarly, the test accuracy of PGD-AT vs GPGD-Attack is 15.53\%, a adversarially-trained model preforms badly on on-manifold attacks. The results on the experiments on CIFAR-10 and ImageNet are similar. PGD-AT is not able to defend GPGD-Attack, and vice versa. In the experiments on CIFAR-10 and ImageNet, we can see that generative attacks have greater attack rates than regular PGD-attacks.

\paragraph{Joint-PGD Adversarial Training} Based on the previous discussion, we may wonder whether we can improve the model robustness by augmenting both on-manifold and regular PGD-adversarial examples to the training set. On MNIST, the jointly-trained model achieves 96.03\% and 95.70\% test accuracy against GPGD-Attack and PGD-Attack, which are similar to the test accuracy of the single-trained models. We can see similar results on CIFAR-10 and ImageNet. Since generative adversarial examples are not closed to the clean data because of the large Lipschitz of the generative models. We study the performance of norm-bounded on-manifold adversarial examples in the next subsection.

\subsection {Eigenspace Adversarial Examples}

\begin{figure*}[htbp]
\centering
\subfigure[]{
\begin{minipage}[t]{0.24\linewidth}
\centering
\includegraphics[width=1.2in]{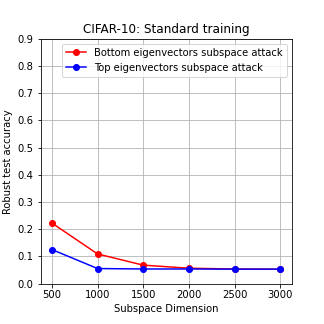}
\end{minipage}%
}%
\subfigure[]{
\begin{minipage}[t]{0.24\linewidth}
\centering
\includegraphics[width=1.2in]{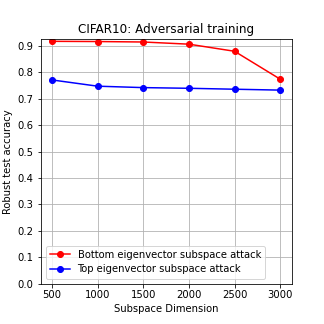}
\end{minipage}%
}%
\subfigure[]{
\begin{minipage}[t]{0.24\linewidth}
\centering
\includegraphics[width=1.2in]{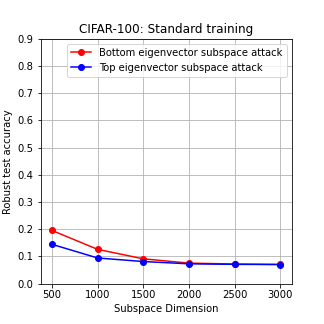}
\end{minipage}
}%
\subfigure[]{
\begin{minipage}[t]{0.24\linewidth}
\centering
\includegraphics[width=1.2in]{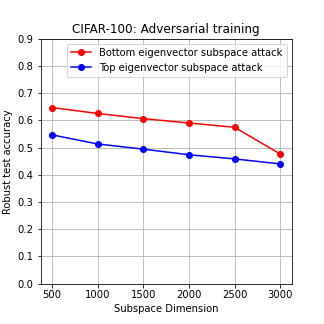}
\end{minipage}
}%
\centering
\caption{Robust test error of standard training and adversarial training against eigenspace attack on CIFAR-10 and CIFAR-100. The x-axis is the dimension of the restricted subspace. The performance against top eigenvectors subspace attacks is shown in the blue lines. The performance against bottom eigenvectors subspace attacks is shown in red lines. (a) Standard-trained model on CIFAR-10. (b) Adversarially-trained model on CIFAR-10. (c) Standard-trained model on CIFAR-100. (d) Standard-trained model on CIFAR-100.}
\label{fig1}
\end{figure*}

\paragraph{Experiment Setups} In Fig. \ref{fig1}, we show the results of the experiments of standard training and $\ell_2$-adversarial training against $\ell_2$-eigenspace adversarial attacks on CIFAR-10 \citep{krizhevsky2009learning} and CIFAR-100. For adversarial training, we use 20 steps PGD-attack with $\varepsilon=0.5$ for the inner maximization problem. For the eigenspace attack, we restrict the subspace in different dimensions. The results of top eigenvectors subspace attacks restricted in the subspace spanned by the first 500, 1000, 1500, 2500, and 3000 eigenvectors are shown in the blue lines. Correspondingly, the results of bottom eigenvectors subspace attacks restricted in the subspace spanned by the last eigenvectors are shown in the red lines.

\paragraph{Target Model} The target model in Fig. \ref{fig1} is Wide-ResNet-28. In the experiments on CIFAR-10, for the standard-trained model, the clean accuracy is 92.14\%. For the adversarial training model, the clean accuracy is 95.71\%, and the robust accuracy against $\ell_2$-PGD-20 is  73.22\%. On CIFAR-100. for standard training, the clean accuracy is 80.34\%. For adversarial training, the clean accuracy is 67.18\%, and the robust accuracy against PGD-10 is 43.92\%.

\paragraph{On-manifold Adversarial Examples Have Greater Attack Rates on Standard-Trained models} In Fig. \ref{fig1} (a), if we restrict the attack directions in the subspace spanned by the first or last 500 eigenvectors, the robust accuracies are 12\% and 22\%. As we increase the dimension of the subspace, the test accuracy will converge to the robust accuracy of standard PGD-attack, 5.27\%. If the dimension of subspace increases to 3072, the subspace attacks are equivalent to standard PGD-attacks since the dimension of $\mathcal{X}$ is $3\times 3\times 32 = 3072$ on CIFAR-10. On-manifold adversarial examples have greater attack rates than off-manifold adversarial examples against the standard training model. The results in Fig. \ref{fig1} (c) are similar.

\paragraph{On-manifold Adversarial Examples Have Greater Attack Rates on Adversarially-Trained model} Now, we turn to Fig. \ref{fig1} (b). If we restrict the attack directions in the first or last 1000 eigenvectors subspace, the robust accuracy is 58.74\% and 66.43\% against bottom and top eigenvectors subspace attacks, respectively. Similar for other numbers of dimensions. Similarly for the experiments on CIFAR-100 in Fig. \ref{fig1} (d). We can see that adversarial training can work well on off-manifold attacks, but the performance on on-manifold attacks is not good enough.

\paragraph{Eigenspace Adversarial Training} In this setting, the perturbation constraint is a subset of the constraint of regular adversarial training. The performance of this algorithm is worse than that of regular adversarial training. We provide the experiments in Appendix B.2.

\paragraph{Adversarial Training Ignores On-manifold Adversarial Examples} In these experiments, all the on-manifold adversarial examples are norm-bounded. Adversarial training still performs badly on this kind of attack. In Theorem \ref{thm:er3} in the next section, we will show that it is because off-manifold adversarial examples have larger losses than on-manifold adversarial examples in the norm constraint. Then, norm-based adversarial training algorithms are not able to find the on-manifold adversarial examples in the small norm ball. Therefore, adversarial training cannot fit the on-manifold adversarial examples well.

\paragraph{\textbf{Take-away Message}} In short, Sec. \ref{sec:perform} shows that (approximate) on-manifold adversarial examples are powerful. It has higher attack rates than off-manifold adversarial examples on both standard-trained and adversarially-trained models. Widely use adversarially-trained models performs badly against (approximate) on-manifold attacks.

\section{Theoretical Analysis: On-manifold Attacks on Gaussian Mixture Model}
\label{sec:ta}
In this section, we study the excess risk and the adversarial distribution shift of on-manifold adversarial training. We study the binary classification setting proposed by \citep{ilyas2019adversarial}. In this setting, the true data manifold is known. We can derive the optimal closed-form solution of on-manifold adversarial attacks and adversarial training and provide insights to understand on-manifold adversarial examples.

\subsection{Theoretical Model Setup}
\paragraph{Gaussian mixture model} Assume that data points $(x,y)$ are sampled according to $y\sim\{-1,1\}$ uniformly and $x\sim\mathcal{N}(y\boldsymbol{\mu_*},\boldsymbol{\Sigma_*})$, where $\bmus$ and $\bsigmas$ denote the true mean and covariance matrix of the data distribution. For the data in the class $y=-1$, we replace $x$ by $-x$, then we can view the whole dataset as sampled from $\mathcal{D}= \mathcal{N}(\boldsymbol{\mu_*},\boldsymbol{\Sigma_*})$.

\paragraph{Classifier} The goal of standard training is to learn the parameters $\boldsymbol{\Theta}=(\boldsymbol{\mu},\boldsymbol{\Sigma})$ such that 
\begin{equation}\label{equ:stdtrain}
\boldsymbol{\Theta}=\arg\min_{\boldsymbol{\mu},\boldsymbol{\Sigma}}\mathcal{L}(\boldsymbol{\mu},\boldsymbol{\Sigma})=\arg\min_{\boldsymbol{\mu},\boldsymbol{\Sigma}}\mathbb{E}_{x\sim\mathcal{D}}[\ell(x;\boldsymbol{\mu},\boldsymbol{\Sigma})],
\end{equation}
 where $\ell(\cdot)$ represents the negative log-likelihood function. The goal of adversarial training is to find 
 \begin{equation}\label{equ:advtrain}
 \begin{aligned}
 \boldsymbol{\Theta_r}=\arg\min_{\boldsymbol{\mu},\boldsymbol{\Sigma}}\mathcal{L}_{r}(\boldsymbol{\mu},\boldsymbol{\Sigma})
 =\arg\min_{\boldsymbol{\mu},\boldsymbol{\Sigma}}\mathbb{E}_{x\sim\mathcal{D}}[\max_{\|x-x'\|\leq\varepsilon}\ell(x';\boldsymbol{\mu},\boldsymbol{\Sigma})].
 \end{aligned}
 \end{equation}
 We use $\mathcal{L}$ and $\mathcal{L}_r$ to denote the standard loss and adversarial loss. After training, we classify a new data point $x$ to the class $\text{sgn}(\boldsymbol{\mu}^T\boldsymbol{\Sigma}^{-1}x)$. 

 \paragraph{Generative Model} In our theoretical study, we use a linear generative model, that is, probabilistic principle components analysis (P-PCA) \citep{tipping1999probabilistic}. P-PCA can be viewed as linear VAE \citep{dai2017hidden}.

 Given dataset $\{x_i\}_{i=1}^n\subset\mathbb{R}^d$, let $\bmu$ and $\boldsymbol{S}$ be the sample mean and sample covariance matrix. The eigenvalue decomposition of $\boldsymbol{S}$ is $\boldsymbol{S}=\bu\blambda \bu^T$, then using the first $q$ eigenvactors, we can project the data to a low dimensional space. P-PCA is to assume that the data are generated by 
\[x=\boldsymbol{W}z+\boldsymbol{\mu}+\boldsymbol{\epsilon}\quad \text{where}\quad z\sim\mathcal{N}(0,I)\ ,\ \boldsymbol{\epsilon}\sim\mathcal{N}(0,\sigma^2I),\]
$z\in\mathbb{R}^q$ and $\bw\in\mathbb{R}^{d\times q}$. Then we have $x\sim\mathcal{N}(\boldsymbol{\mu},\boldsymbol{W}\boldsymbol{W}^T+\sigma^2I)$, $x|z\sim\mathcal{N}(\boldsymbol{W}z+\boldsymbol{\mu},\sigma^2I)$ and $z|x\sim\mathcal{N}(\boldsymbol{P}^{-1}\boldsymbol{W}^T(x-\boldsymbol{\mu}),\sigma^2\boldsymbol{P}^{-1})$ where $\boldsymbol{P}=\boldsymbol{W}^T\boldsymbol{W}+\sigma^2I$. The maximum likelihood estimator of $\boldsymbol{W}$ and $\sigma^2$ are
\[
\boldsymbol{W}_{\text{ML}}=\boldsymbol{U}_q(\boldsymbol{\Lambda}_q-\sigma_{\text{ML}}^2I)^{1/2}\ \text{and}\ \sigma^2_{\text{ML}}=\frac{1}{d-q}\sum_{i=q+1}^d\lambda_i,
\]
where $\buq$ is the matrix of the first $q$ columns of $\bu$, $\blambdaq$ is the matrix of the first $q$ eigenvalues of $\blambda$. In the following study, we assume that $n$ is large enough such that we can learn the true $\boldsymbol{\mu}_*$ and $\bsigmas$. Thus we have $\boldsymbol{S}=\boldsymbol{\Sigma}_*$, $\boldsymbol{U}_q=\boldsymbol{U}_{q*}$, $\boldsymbol{\Lambda}_q=\boldsymbol{\Lambda}_{q*}$ for the generative model. 

\paragraph{Generative Adversarial Examples}
To perturb the data in the latent space, data will go through the encode-decode process $x\rightarrow z\rightarrow\Delta z+z\rightarrow x'$. Specifically,  we sample $x\sim\mathcal{D}$, then we encode $z=\arg\max q(z|x)=\boldsymbol{P}^{-1}\boldsymbol{W}^T(x-\boldsymbol{\mu_*})$, add a perturbation $\Delta z$, and finally, we decode $x_{adv}=\arg\max p(x|z+\Delta z)=\boldsymbol{W}(z+\Delta z)+\boldsymbol{\mu_*}$. The following lemma shows that the generative adversarial examples can be rewritten as a  new form.

\begin{lemma}[Generative Adversarial Examples]
\label{lemma1}
Using generative model P-PCA, the adversarial examples can be rewritten as
\begin{equation*}
\begin{aligned}
x_{adv}=x'+\boldsymbol{W}\Delta z\quad& \text{and} \quad x'\sim\mathcal{D}'=\mathcal{N}(\boldsymbol{\mu_*},\boldsymbol{U_*}\boldsymbol{\Lambda}'\boldsymbol{U_*}^T),\\ \text{with}\ \blambda'=&\begin{bmatrix}
(\blambdaq-\sigma^2I)^2\blambdaq^{-1}& 0\\
0&0
\end{bmatrix}.
\end{aligned}
\end{equation*}
If the data lie in a $q$ dimensional subspace, i.e. the covariance matrix $\bsigmas$ is rank $q$, we have $\blambda'=\blambdas$. Then $\mathcal{D}'=\mathcal{D}$.
\end{lemma}

\paragraph{Remark} We consider other sampling strategies using generative models in \ref{A}. We show that the adversarial examples have the same form as in Lemma \ref{lemma1} with different $\blambda'$.

The adversarial expected risk of \emph{generative adversarial training} can be rewritten as the following minimax problem
\begin{equation}\label{equ:ls2}
\begin{aligned}
\min_{\boldsymbol{\mu},\boldsymbol{\Sigma}}\mathcal{L}_{gat}(\boldsymbol{\mu},\boldsymbol{\Sigma};\mathcal{D}')
=\min_{\boldsymbol{\mu},\boldsymbol{\Sigma}}\mathbb{E}_{x'\sim\mathcal{D}'}\max_{\|\Delta z\|\leq\varepsilon}\ell(x'+\boldsymbol{W}\Delta z,\boldsymbol{\mu},\boldsymbol{\Sigma}).
\end{aligned}
\end{equation}

\paragraph{Eigenspace Adversarial Examples} If the perturbation is restricted in the eigenspace spanned by $U_q$. the adversarial examples can be weritten as $x_{adv}=x+\boldsymbol{U}_q \Delta z$. 

The adversarial expected risk of \emph{eigenspace adversarial training} can be rewritten as the following minimax problem
\begin{equation}\label{equ:ls3}
\begin{aligned}
\min_{\boldsymbol{\mu},\boldsymbol{\Sigma}}\mathcal{L}_{eat}(\boldsymbol{\mu},\boldsymbol{\Sigma};\mathcal{D})
=\min_{\boldsymbol{\mu},\boldsymbol{\Sigma}}\mathbb{E}_{x\sim\mathcal{D}}\max_{\|\Delta z\|\leq\varepsilon}\ell(x+\boldsymbol{U}_q\Delta z,\boldsymbol{\mu},\boldsymbol{\Sigma}).
\end{aligned}
\end{equation}
When $q=d$, $\boldsymbol{U}_q$ span the whole space. Eq. (\ref{equ:ls3}) reduce to the regular adversarial training.

\subsection{Excess Risk Analysis}
We consider the excess risk incurred by the optimal perturbation of on-manifold adversarial examples, i.e. $\mathcal{L}_{gat}-\mathcal{L}$ and $\mathcal{L}_{eat}-\mathcal{L}$ given the true $\boldsymbol{\Theta_*}$.

We consider the Lagrange penalty form of the inner maximization problem in Eq. (\ref{equ:ls2}), i.e.  $\max\ell(x'+\bw\Delta z,\bmu,\bsigma)-L\|\Delta z\|^2/2$, where $L$ is the Lagrange multiplier. Similarly, we consider the Lagrange penalty form of the inner maximization problem in Eq. (\ref{equ:ls3}).

\begin{theorem}[Excess risk of generative adversarial examples]
\label{thm:er2}
Let $\mathcal{L}_{gat}$ and $\mathcal{L}$ be the loss in equations (\ref{equ:ls2}) and (\ref{equ:stdtrain}) respectively), given the non-robustly learned $\boldsymbol{\Theta_*}=(\boldsymbol{\mu_*},\boldsymbol{\Sigma_*})$, the excess risk is
\begin{equation*}
\begin{aligned}
&\mathcal{L}_{gat}(\boldsymbol{\Theta_*},\mathcal{D}')-\mathcal{L}(\boldsymbol{\Theta_*},\mathcal{D})\\
=&\frac{1}{2}\sum_{i=1}^q\Big[(1+\frac{\lambda_i-\sigma^2}{(L-1)\lambda_i+\sigma^2})^2-1\Big]\frac{\lambda_i'}{\lambda_i}
+\frac{1}{2}\log\Big[\frac{\prod_{i=1}^{d}\lambda_i'}{\prod_{i=1}^{d}\lambda_i}\Big]+\frac{1}{2}\big(\sum_{i=1}^d\frac{\lambda_i'}{\lambda_i}-d\big).
\end{aligned}
\end{equation*}
If the data lie in a low dimensional manifold, $\text{rank}(\bsigmas)=q$, the excess risk is
	\[
	\mathcal{L}_{gat}(\boldsymbol{\Theta_*},\mathcal{D}')-\mathcal{L}(\boldsymbol{\Theta_*},\mathcal{D})=\Theta(qL^{-2}).
	\]
\end{theorem}

The optimal perturbation in the latent space will incur an excess risk in $\Theta(qL^{-2})$. The adversarial vulnerability depends on the dimension $q$ and the Lagrange multiplier $L$. $L$ is negatively related to the perturbation intensity $\varepsilon$. Then, we analyze the excess risk of eigenspace adversarial examples. Since the perturbation thresholds, $\varepsilon$, are on different scales, the corresponding Lagrange multipliers $L$ are different. We use $L_2$ in the following Theorem.

\begin{theorem}[Excess risk of eigenspace adversarial examples]\label{thm:er3}
	Let $\mathcal{L}_{eat}$ and $\mathcal{L}$ be the loss in equations (\ref{equ:ls3}) and (\ref{equ:stdtrain}) respectively, given the non-robustly learned $\boldsymbol{\Theta_*}=(\boldsymbol{\mu_*},\boldsymbol{\Sigma_*})$, denote $\lambda_{q}$ be the $q^{th}$ eigenvalue of $\bsigmas$, the excess risk satisfies
	\begin{equation*}\begin{aligned}
	&\mathcal{L}_{eat}(\boldsymbol{\Theta_*},\mathcal{D})-\mathcal{L}(\boldsymbol{\Theta_*},\mathcal{D})
	\leq \mathcal{O}(q(\lambda_{q}L_2)^{-2})\ \text{and}\\
	&\mathcal{L}_{eat}(\boldsymbol{\Theta_*},\mathcal{D})-\mathcal{L}(\boldsymbol{\Theta_*},\mathcal{D})\geq\Omega((\lambda_{q}L_2)^{-2}).
	\end{aligned}\end{equation*}
\end{theorem} 

Similar to the case of generative attacks, the adversarial vulnerability depends on the dimension $q$ and the Lagrange multiplier $L2$.

\paragraph{Excess risk of (regular) adversarial examples} When $q=d$ and the data lie in a low dimensional manifold, i.e. $\lambda_{min}=0$, the excess risk of regular adversarial examples equals to $+\infty$. Regular adversarial attacks focus on the directions corresponding to zero eigenvalues, i.e. the off-manifold directions. However, on-manifold adversarial examples will also incur large excess risks in $\Theta(qL^{-2})$ and $\Omega((\lambda_{q}L_2)^{-2})$. On-manifold adversarial examples are also powerful but regular adversarial attacks tend to ignore the on-manifold adversarial examples.

 \begin{figure*}[ht]
 	\centering
 	\includegraphics[width=\linewidth]{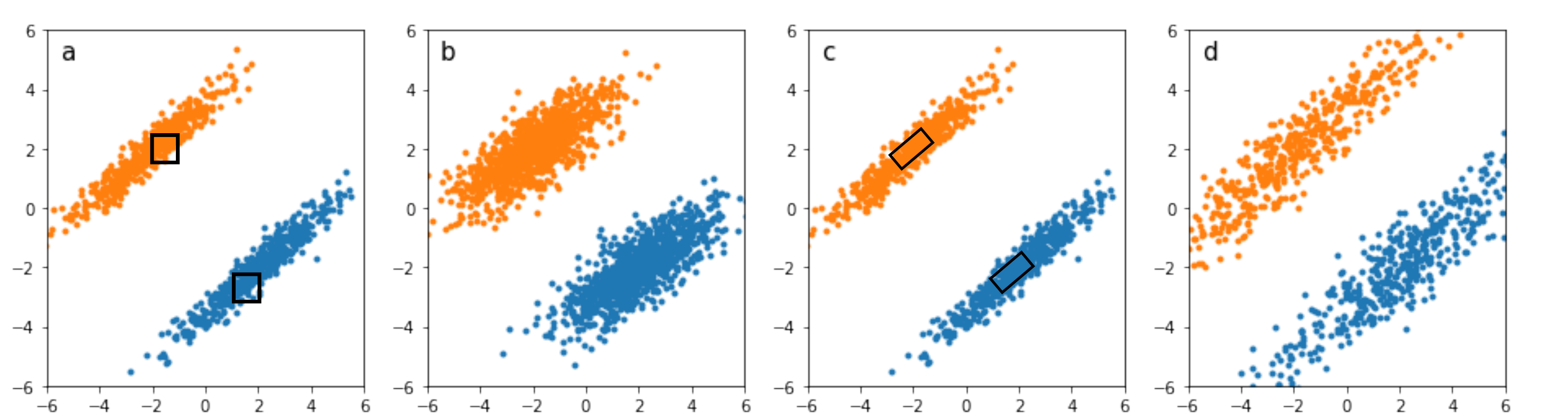}
  	\vspace{-0.2in}
 	\caption{Demonstration and numerical simulation of theoretical analysis. Let the ellipses be the Gaussian data, and let the black block be the perturbation constraint with the original example in the center. (a) The constraint of eigenspace adversarial attacks; (b) Adversarial distribution shift of eigenspace adversarial training; (c)  The constraint of generative adversarial attacks; (d) Adversarial distribution shifts of generative adversarial training.}
 	\label{fig2}
 \end{figure*}

\subsection{Adversarial Distribution Shifts}
 In this subsection we study the optimal solution of optimization problem in Eq.  (\ref{equ:ls2}) and (\ref{equ:ls3}). Since they are not standard minimax problems, we consider a modified problem of (\ref{equ:ls2}):
 \begin{equation}
 \label{equ:modify}
 \min_{\boldsymbol{\mu},\boldsymbol{\Lambda}}\max_{\mathbb{E}_{x'}\|\Delta z\|=\varepsilon}\mathbb{E}_{x'\sim\mathcal{D}'}\ell(x'+\boldsymbol{W}\Delta z,\boldsymbol{\mu},\boldsymbol{\Sigma}).
 \end{equation}
 
  Similarly, we switch the order of expectation and maximization in problem (\ref{equ:ls3}). The following two Theorems are our main results. They give the optimal solution of generative adversarial training and eigenspace adversarial training.

\begin{theorem}[Optimal solution of generative adversarial training]\label{thm:alp2}
	The optimal solution of the modified problem in Eq. (\ref{equ:modify}) is
	\[\boldsymbol{\mu}_{eat}=\boldsymbol{\mu_*}\quad \text{and} \quad
	\boldsymbol{\Sigma}_{eat}=\boldsymbol{U_*}\boldsymbol{\Lambda}^{gat}\boldsymbol{U_*}^T,\ \ \text{	where}
	\]
	\begin{equation*}
	\begin{aligned} &\lambda_i^{gat}=\frac{1}{4}\Big[2\lambda_i'+\frac{4(\lambda_i-\sigma^2)}{L}+2\lambda_i'\sqrt{1+\frac{4(\lambda_i-\sigma^2)}{\lambda_i'L}}\Big],\\
	 &\text{for}\ i=1\leq q,\ \ \text{and}\ 
	\lambda_i^{gat}=\lambda_i'\ \ \text{for}\ \ i>q.
	\end{aligned}
	\end{equation*}
\end{theorem}

We assume that the data lie in a $q$-dimensional manifold again. Then we have $\lambda_i^{ls}/\lambda_i=1/2+1/L+\sqrt{1/4+1/L}\geq 1$ for $i\leq q$ and $\lambda_i^{ls}/\lambda_i=0$ for $i>q$. Generative adversarial training amplifies the largest $q$ eigenvalues of the covariance matrix of the (adversarial) data distribution.

\begin{theorem}[Optimal solution of eigenspace adversarial training]
	The optimal solution of the modified problem of eigenspace adversarial training\footnote{Remark: Thm. \ref{thm:er3} and Thm. \ref{thm:robust} provide the excess risk and adversarial distribution shift of regular adversarial training by setting $q=d$.} is
	\[\boldsymbol{\mu}_{eat}=\boldsymbol{\mu_*}\quad \text{and} \quad
	\boldsymbol{\Sigma}_{eat}=\boldsymbol{U_*}\boldsymbol{\Lambda}^{eat}\boldsymbol{U_*}^T,\ \ \text{where}
	\]
	\begin{equation*}
	\begin{aligned} &\lambda_i^{eat}=\frac{1}{4}\Big[2\lambda_i+\frac{4}{L_2}+2\lambda_i\sqrt{1+\frac{4}{\lambda_iL_2}}\Big],\\
	 &\text{for}\ i=1\leq q,\ \ \text{and}\ 
	\lambda_i^{eat}=\lambda_i\ \ \text{for}\ \ i>q.
	\end{aligned}
	\end{equation*}
	\label{thm:robust}
\end{theorem}
Considering the ratio
\[
\frac{\lambda_i^{eat}}{\lambda_i}=\frac{1}{2}+\frac{1}{L_2\lambda_i}+\sqrt{\frac{1}{L_2\lambda_i}+\frac{1}{4}},
\]
the ratio $\lambda_i^{eat}/\lambda_i$ is larger for smaller true eigenvalue $\lambda_i$. Therefore, eigenspace adversarial training amplifies the small eigenvalues of the eigenspace. 

\paragraph{Optimal solution of (regular adversarial training)} When $q=d$, regular adversarial training amplifies the small eigenvalues of the whole space.

\paragraph{Demonstration of the case $q=2$} We give a numerical simulation in Fig. \ref{fig2}. The black blocks are the perturbation constraint with the original example in the center. Given the original Gaussian data (plots in (a)), we iteratively solve the max and min problem in Eq. (\ref{equ:ls3}), the resulting data distribution is plotted in (b). We can see that the radius of the short axis of the ellipse is amplified. In Fig. \ref{fig2} (d), we iteratively solve the max and min problem in Eq. (\ref{equ:ls2}), we can see that the radius of the two axes of the ellipse is amplified.

\paragraph{\textbf{Take-away Message}} In the analysis of GMMs, we prove that regular adversarial attacks and defense only focus on off-manifold directions due to the magnitude of eigenvalues. However, on-manifold adversarial examples also incur large excess risk and lead to different adversarial distribution shift. This is why (regular) adversarial training performs badly against on-manifold attacks. 
\section{Further Analysis of On-manifold Attacks}
\label{sec:exper}
\begin{figure}
	\vskip -0.23in
	\centering
	\scalebox{1}{
		\subfigure[]{
			\begin{minipage}[htp]{0.233\textwidth}
				\centering
				\includegraphics[width=1\textwidth]{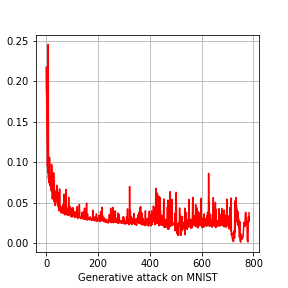}
			\end{minipage}
		}
		\subfigure[]{
			\begin{minipage}[htp]{0.233\textwidth}
				\centering
				\includegraphics[width=1\textwidth]{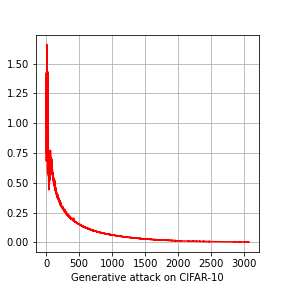}
			\end{minipage}
	}}
	\caption{(a) The average of the generative attack directions on MNIST. (b) The average of the generative attack directions on CIFAR-10.}
		\vskip -0.23in
	\label{genattack}
\end{figure}
We provide further analysis of the theoretical properties on real dataset. We will see that the properties derived in Sec. \ref{sec:ta} can also be observed in practice.

\paragraph{On-manifold Attack Directions} Given the training dataset $\mathcal{D}$, we conduct SVD on this dataset and let the eigenvectors span a basis of the original space. In Fig. \ref{genattack} (a) and (b), we plot the absolute value of the average of all the generative adversarial examples in this space. We can see that generative attack focuses on the directions of top eigenvectors. Gen-AE and Eigen-AE have similar attack directions on real datasets.

\begin{figure}[t]
	\centering
	\scalebox{0.9}{
			\begin{minipage}[htp]{0.01\linewidth}
			\centering
			\rotatebox{90}{\text{MNIST}}
		\end{minipage}
		\subfigure[]{
			\begin{minipage}[htp]{0.23\linewidth}
				\centering
				\includegraphics[width=1.3in]{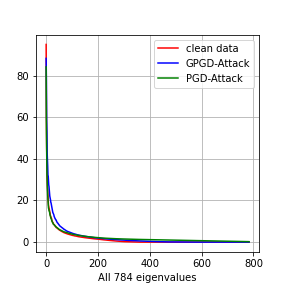}
			\end{minipage}%
		}
		\subfigure[]{
			\begin{minipage}[htp]{0.23\linewidth}
				\centering
				\includegraphics[width=1.3in]{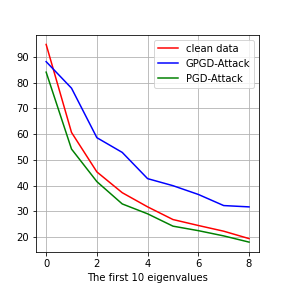}
			\end{minipage}
		}
		\subfigure[]{
			\begin{minipage}[htp]{0.23\linewidth}
				\centering
				\includegraphics[width=1.3in]{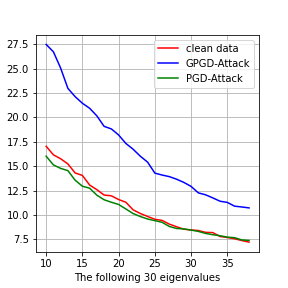}
			\end{minipage}
		}
		\subfigure[]{
			\begin{minipage}[htp]{0.23\linewidth}
				\centering
				\includegraphics[width=1.3in]{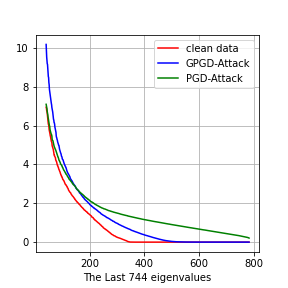}
			\end{minipage}
	}}
\scalebox{0.9}{

		\begin{minipage}[htp]{0.01\linewidth}
			\centering
			\rotatebox{90}{\text{CIFAR-10}}
		\end{minipage}%
	\subfigure[]{
		\begin{minipage}[htp]{0.23\linewidth}
			\centering
			\includegraphics[width=1.3in]{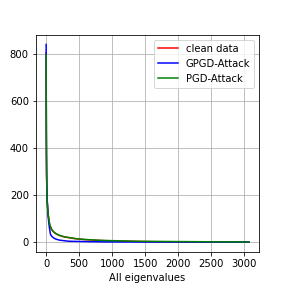}
		\end{minipage}%
	}
	\subfigure[]{
		\begin{minipage}[htp]{0.23\linewidth}
			\centering
			\includegraphics[width=1.3in]{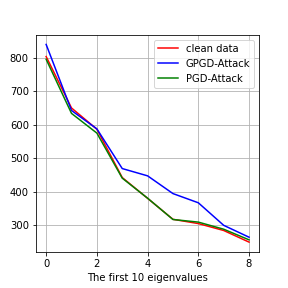}
		\end{minipage}
	}
	\subfigure[]{
		\begin{minipage}[htp]{0.23\linewidth}
			\centering
			\includegraphics[width=1.3in]{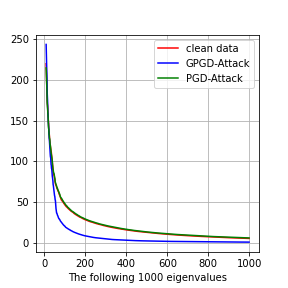}
		\end{minipage}
	}
	\subfigure[]{
		\begin{minipage}[htp]{0.23\linewidth}
			\centering
			\includegraphics[width=1.3in]{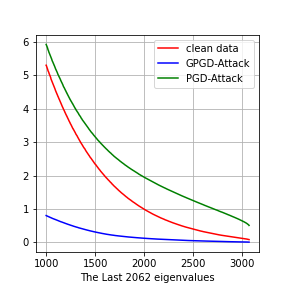}
		\end{minipage}
}}
	\centering
	\caption{Adversarial distribution shifts on MNIST and CIFAR-10. The first row is the experiment on MNIST. The second row is the experiment on CIFAR-10. (a) and (e) plot all the eigenvalues of the dataset. (b) and (f): generative adversarial training (blue line) will amplify the large eigenvalue of the dataset, while adversarial training will not. (d) and (h) Adversarial training (green line) will amplify the small eigenvalues of the dataset, while generative adversarial training will not.}
	\label{eigens}
\end{figure}

\paragraph{Adversarial Distribution Shifts} Let $f_{at}(x)$ and $f_{gat}(x)$ be the adversarially trained and generative adversarially trained neural networks respectively. Given the clean dataset $\mathcal{D}$, let the robust dataset be $$\mathcal{D}_{at}=\{x_{adv}| x_{adv} = \arg\max\ell(f_{at}(x),y),(x,y)\in\mathcal{D}\}$$ and $\mathcal{D}_{gat}=\{x_{adv}| x_{adv} = \arg\max\ell(f_{gat}(G(z)),y),(x,y)\in\mathcal{D}\}$. We perform SVD on $\mathcal{D}$, $\mathcal{D}_{at}$, and $\mathcal{D}_{gat}$. We plot the eigenvalues in Fig. \ref{eigens}. The first row is the experiments on MNIST,  and the second row is the experiments on CIFAR-10. On MNIST, the number of dimensions is $1\times 28\times28=784$. On CIFAR-10, the numbers of dimension is $3\times 32\times 32=3072$.

\paragraph{Generative adversarial training amplifies the top eigenvalues} In Fig. \ref{eigens} (b) and (f), we plot the top 10 eigenvalues. we can see that the eigenvalues of $\mathcal{D}_{gat}$ (blue line) is larger than that of $\mathcal{D}$ (red line). It means that generative adversarial training will amplify the large eigenvalues, which is consistent with Theorem \ref{thm:alp2}. Regular adversarial training does not amplify the top eigenvalues.

\paragraph{Adversarial training amplifies the bottom eigenvalues} In Fig. \ref{eigens} (d) and (h), we can see that the bottom eigenvalues of $\mathcal{D}_{at}$ (green line) is larger than that of $\mathcal{D}$ (red line), which is consistent with Theorem \ref{thm:robust}, while the bottom eigenvalues of $\mathcal{D}_{gat}$ is not larger than that of $\mathcal{D}$. Therefore, Adversarial training amplify the bottom eigenvalues.

\paragraph{Discussion on Adversarial distribution shifts} In general, it is hard to characterize the adversarial distribution shifts in detail since analyzing the minimax problem of adversarial training is challenging. In these experiments, we show that the eigenvalues, which are second-order statistics of the dataset or the variance of the data distribution in the corresponding directions, are good indicators of the adversarial distribution shifts. Adversarial training tries to fit the adversarial examples in directions of small variance, while generative adversarial training tries to fit the adversarial examples in directions of large variance. Therefore, adversarial training ignores on-manifold adversarial examples, which are also important for adversarial robustness.

\section{Conclusion}
\label{future}
In this paper, We show that on-manifold adversarial examples are also powerful, but adversarial training focuses on off-manifold directions and ignores the on-manifold adversarial examples. On-manifold adversarial examples are also important for adversarial training. We think that our analysis can inspire more theoretical research on adversarial robustness. For example, A well-designed norm-bounded on-manifold adversarial attacks algorithm may improve the performance of adversarial training.

\clearpage
\bibliographystyle{apalike}
\bibliography{main.bib}

\clearpage
\appendix
\section{Proof of the Theorems}
\label{A}
\subsection{Proof of Lemma \ref{lemma1}}
We first introduce the general form of Lemma \ref{lemma1}. To perturb the data in the latent space, data will go through the encode-decode process $x\rightarrow z\rightarrow\Delta z+z\rightarrow x'$. Based on the probabilistic model, we may choose $z$ with the highest probability or just sample it from the distribution we learned. Hence, we could have different strategies.

Strategy 1: Sample $x\sim\mathcal{D}$, encode $z=\arg\max q(z|x)=\boldsymbol{P}^{-1}\boldsymbol{W}^T(x-\boldsymbol{\mu_*})$, add a perturbation $\Delta z$, and finally, decode $x_{adv}=\arg\max p(x|z+\Delta z)=\boldsymbol{W}(z+\Delta z)+\boldsymbol{\mu_*}$.

Strategy 2: Sample $x\sim\mathcal{D}$, then sample $z\sim q(z|x)$, add a perturbation $\Delta z$, and finally, sample $x_{adv}\sim p(x|z+\Delta z)$.

Strategy 3: Sample $z\sim\mathcal{N}(0,I)$, add a perturbation $\Delta z$, and then sample $x_{adv}\sim p(x|z+\Delta z)$. In this strategy, $x_{adv}$ can be viewed as the adversarial example of $x=\arg\max_x q(z|x)$.

Lemma \ref{lemmaA1} below is a general form of the Lemma \ref{lemma1}, where we only consider strategy 1 ($j=1$). Similar to the proof of Thm. \ref{thm:er2} to \ref{thm:robust}.
\begin{lemma}[generative adversarial examples]
	\label{lemmaA1}
	Using the 3 strategies defined above, the generative adversarial examples can be unified as
	\begin{equation*}
	x_{adv}=x'+\boldsymbol{W}\Delta z\quad \text{and}\quad x'\sim\mathcal{D}'_j=\mathcal{N}(\boldsymbol{\mu_*},\boldsymbol{U_*}\boldsymbol{\Lambda}^{(j)}\boldsymbol{U_*}^T), \quad j=1,2,3,
	\end{equation*}
	where
	\[
	\blambda^{(1)}=\begin{bmatrix}
	(\blambdaq-\sigma^2I)^2\blambdaq^{-1}& 0\\
	0&0
	\end{bmatrix}\quad \blambda^{(3)}=\begin{bmatrix}
	\blambdaq&0\\
	0&\sigma^2I
	\end{bmatrix}
	\]
	\[
	\blambda^{(2)}=\begin{bmatrix}
	(\blambdaq-\sigma^2I)^2\blambdaq^{-1}+(\blambdaq-\sigma^2I)\blambdaq^{-1}\sigma^2+\sigma^2I& 0\\
	0&\sigma^2I
	\end{bmatrix}.
	\]
	If the data lie in a $q$ dimensional subspace, i.e. the covariance matrix $\bsigmas$ is rank $q$, we have $\blambda^{(1)}=\blambda^{(2)}=\blambda^{(3)}=\blambdas$. Then $\mathcal{D}'=\mathcal{D}$.
\end{lemma}

Proof:

Using P-PCA generative model, $x\sim\mathcal{N}(\boldsymbol{\mu},\boldsymbol{W}\boldsymbol{W}^T+\sigma^2I)$, $x|z\sim\mathcal{N}(\boldsymbol{W}z+\boldsymbol{\mu},\sigma^2I)$ and $z|x\sim\mathcal{N}(\boldsymbol{P}^{-1}\boldsymbol{W}^T(x-\boldsymbol{\mu}),\sigma^2\boldsymbol{P}^{-1})$ where $\boldsymbol{P}=\boldsymbol{W}^T\boldsymbol{W}+\sigma^2I$. The maximum likelihood estimator of $\boldsymbol{W}$ and $\sigma^2$ are
\[
\boldsymbol{W}_{\text{ML}}=\boldsymbol{U}_q(\boldsymbol{\Lambda}_q-\sigma^2I)^{1/2}\quad\text{and}\quad \sigma^2_{\text{ML}}=\frac{1}{d-q}\sum_{i=q+1}^d\lambda_i.
\]

\paragraph{Strategy 1} Sample $x\sim\mathcal{D}$, encode $z=\arg\max q(z|x)=\boldsymbol{P}^{-1}\boldsymbol{W}^T(x-\boldsymbol{\mu_*})$, add a perturbation $\Delta z$, and finally, decode $x_{adv}=\arg\max p(x|z+\Delta z)=\boldsymbol{W}(z+\Delta z)+\boldsymbol{\mu_*}$.
Then
\begin{equation*}
\begin{aligned}
x_{adv}=&\boldsymbol{W}(\bp\bw^T(x-\bmus)+\Delta z)+\boldsymbol{\mu_*}\\
=&\boldsymbol{W}\bp\bw^T(x-\bmus)+\boldsymbol{\mu_*}+\bw\Delta z\\
=&x'+\bw\Delta z.\\
\end{aligned}
\end{equation*}
Since $x\sim\mathcal(\bmus,\bsigmas)$, we have $x-\bmus\sim\mathcal(0,\bsigmas)$, Then
\[
x'\sim\mathcal{N}(\bmus,\bw\bp\bw^T\bsigmas(\bw\bp\bw^T)^{T}),
\]
With
\begin{equation*}
\begin{aligned}
&\bw\bp\bw^T\bsigmas(\bw\bp\bw^T)^{T}\\
=&\bus\begin{bmatrix}
\blambdaq-\sigma^2&0\\
0&0
\end{bmatrix}^{1/2}
\begin{bmatrix}
\blambdaq&0\\
0&\sigma^2I
\end{bmatrix}^{-1}
\begin{bmatrix}
\blambdaq-\sigma^2&0\\
0&0
\end{bmatrix}^{1/2}\blambdas\\
&\begin{bmatrix}
\blambdaq-\sigma^2&0\\
0&0
\end{bmatrix}^{1/2}
\begin{bmatrix}
\blambdaq&0\\
0&\sigma^2I
\end{bmatrix}^{-1}
\begin{bmatrix}
\blambdaq-\sigma^2&0\\
0&0
\end{bmatrix}^{1/2}\bus^T\\
=&\bus\begin{bmatrix}
(\blambdaq-\sigma^2I)^2\blambdaq^{-1}& 0\\
0&0
\end{bmatrix}\bus^T\\
=&\bus\blambdaj\bus^T,\quad j=1.
\end{aligned}
\end{equation*}
\paragraph{Strategy 2} Sample $x\sim\mathcal{D}$, then sample $z\sim q(z|x)$, add a perturbation $\Delta z$, and finally, sample $x_{adv}\sim p(x|z+\Delta z)$. Then
\[
z\sim\mathcal{N}(0,\bp\bw^T\bsigmas(\bp\bw^T)^T+\sigma^2\bp)
\]
and
\[
x_{adv}\sim\mathcal{N}(\bmus+\bw\Delta z,\bw\bp\bw^T\bsigmas(\bp\bw^T)^T\bw^T+\bw\sigma^2\bp\bw^T+\sigma^2I),
\]
\[
x_{adv}=x'+\bw\Delta z,
\]
With
\begin{equation*}
\begin{aligned}
&\bw\bp\bw^T\bsigmas(\bp\bw^T)^T\bw^T+\bw\sigma^2\bp\bw^T+\sigma^2I\\
=& \bus\begin{bmatrix}
(\blambdaq-\sigma^2I)^2\blambdaq^{-1}& 0\\
0&0
\end{bmatrix}\bus^T+\bus\begin{bmatrix}
(\blambdaq-\sigma^2I)\blambdaq^{-1}\sigma^2&0\\
0&0
\end{bmatrix}\bus^T+\sigma^2I\\
=& \bus\begin{bmatrix}
(\blambdaq-\sigma^2I)^2\blambdaq^{-1}+(\blambdaq-\sigma^2I)\blambdaq^{-1}\sigma^2+\sigma^2I& 0\\
0&\sigma^2I
\end{bmatrix}\bus^T\\
=&\bus\blambdaj\bus^T,\quad j=2.
\end{aligned}
\end{equation*}
\paragraph{Strategy 3} Sample $z\sim\mathcal{N}(0,I)$, add a perturbation $\Delta z$, and then sample $x_{adv}\sim p(x|z+\Delta z)$. In this strategy, $x_{adv}$ can be viewed as the adversarial example of $x=\arg\max_x q(z|x)$.
\[
x_{adv}\sim\mathcal{N}(\bmus+\bw\Delta z, \bw\bw^T+\sigma^2I),
\]
With
\begin{equation*}
\begin{aligned}
&\bw\bw^T+\sigma^2I\\
=&\bus\begin{bmatrix}
\blambdaq&0\\
0&\sigma^2I
\end{bmatrix}\bus^T\\
=&\bus\blambdaj\bus^T,\quad j=3.
\end{aligned}
\end{equation*}
In these 3 strageties, the adversarial examples can be summerized as
\begin{equation*}
x_{adv}=x'+\boldsymbol{W}\Delta z\quad \text{and}\quad x'\sim\mathcal{D}'_j,\quad j=1,2,3,
\end{equation*}
where $j=1,2,3$ corresponding to strategy 1,2 and 3.

If the data lie in a low dimensional space, i.e. the covariance matrix $\bsigmas$ is rank $q$. Then the maximum likelihood of $\sigma^2_{ML}=\sum_{i=q+1}^d\lambda_i/(d-q)=0$. Then
\[
\blambda^{(1)}=\blambda^{(2)}=\blambda^{(3)}=\begin{bmatrix}
\blambdaq& 0\\
0& 0
\end{bmatrix}=\blambdas.
\]
\qed

\subsection{Proof of Thm. \ref{thm:er2}}
Before we prove Thm. \ref{thm:er2}. We need to prove the following lemma first.
\begin{lemma}[optimal perturbation]
\label{lemma:optperb}
	Given $\boldsymbol{\Theta}=(\boldsymbol{\mu},\boldsymbol{\Sigma})$, the optimal solution of the inner max problem in equation \ref{equ:ls2} is
	\[
	\Delta z^*=\boldsymbol{W}^T(L\boldsymbol{\Sigma}-\boldsymbol{W}\boldsymbol{W}^T)^{-1}(x'-\boldsymbol{\mu}),
	\]
	where $L$ is the lagrange multiplier satisfying $\|\Delta z^*\|=\varepsilon$.
\end{lemma}
Proof: Consider problem
\[
\max_{\|\Delta z\|\leq\varepsilon}\ell(x'+\boldsymbol{W}\Delta z,\boldsymbol{\mu},\boldsymbol{\Sigma}).
\]
The Lagrangian function is
\begin{equation*}
\begin{aligned}
&\ell(x'+\boldsymbol{W}\Delta z,\boldsymbol{\mu},\boldsymbol{\Sigma})-\frac{L}{2}(\|\Delta z\|^2-\varepsilon^2)\\
= &\frac{d}{2}\log(2\pi)+\frac{1}{2}\log|\bsigma|+\frac{1}{2}(x'-\bmu+\bw\Delta z)^T\bsigma^{-1}(x'-\bmu+\bw\Delta z)\\
&-\frac{L}{2}(\|\Delta z\|^2-\varepsilon^2).
\end{aligned}
\end{equation*}
Notice that this quadratic objective function is concave when $L$ is larger than the largest eigenvalue of $\bw^T\bsigma^{-1}\bw$. Calculate the partial derivative with respect to $\Delta z$ and set it to be zero, we have
\begin{equation*}
\begin{aligned}
&\bw^T\bsigma^{-1}(x'-\bmu+\bw\Delta z^*)-L\Delta z^*=0\\
\Leftrightarrow&(L-\bw^T\bsigma^{-1}\bw)\Delta z^* =\bw^T\bsigma^{-1}(x'-\bmu)\\
\Leftrightarrow&\Delta z^* =(L-\bw^T\bsigma^{-1}\bw)^{-1}\bw^T\bsigma^{-1}(x'-\bmu)\\
\Leftrightarrow & \Delta z^*= \bw^T(L\bsigma-\bw\bw^T)^{-1}(x'-\bmu).
\end{aligned}
\end{equation*}
The last equation comes from the Woodbury matrix inversion Lemma. We can obtain $L$ by solving the equation $\|\Delta z^*\|=\varepsilon$.\qed

Now we move to the proof of Thm. \ref{thm:er2}. To derive the expression of excess risk, we decompose it into two parts

\begin{equation}
\begin{aligned}
&\mathcal{L}_{gat}(\boldsymbol{\Theta_*};\mathcal{D}_j')-\mathcal{L}(\boldsymbol{\Theta_*};\mathcal{D})\\
=&\underbrace{\mathcal{L}_{gat}(\boldsymbol{\Theta_*};\mathcal{D}_j')-\mathcal{L}(\boldsymbol{\Theta_*};\mathcal{D}_j')}_{perturbation}+\underbrace{\mathcal{L}(\boldsymbol{\Theta_*};\mathcal{D}_j')-\mathcal{L}(\boldsymbol{\Theta_*};\mathcal{D})}_{change\ of\ distribution}.
\end{aligned}
\end{equation}
Since \[x'\sim\mathcal{D}_j=\mathcal{N}(\bmus,\bsigmaj)=\mathcal{N}(\bmus,\bus\blambdaj\bus^T).\]
We denote $v$ as
\[v=x'-\bmus\sim\mathcal{N}(0,\bus\blambdaj\bus^T).\]
Besides, we have
\begin{equation*}
\begin{aligned}
&\bw\bw^T=\buq(\blambdaq-\sigma^2I)\buq^T
=\bus\begin{bmatrix}
\blambdaq-\sigma^2I& 0\\
0& 0
\end{bmatrix}\bus^T.
\end{aligned}
\end{equation*}
Then, the excess risk caused by perturbation is
\begin{equation*}
\begin{aligned}
&2(\mathcal{L}_{gat}(\boldsymbol{\Theta_*},\mathcal{D}_j')-\mathcal{L}(\boldsymbol{\Theta_*},\mathcal{D}_j'))\\
= &\mathbb{E}(v+\bw\bw^T(L\bsigmas-\bw\bw^T)^{-1}v)^T\bsigmasi (v+\bw\bw^T(L\bsigmas-\bw\bw^T)^{-1}v)-\mathbb{E} v^T\bsigmasi v\\
=&Tr\big[
(I+\bw\bw^T(L\bsigmas-\bw\bw^T)^{-1})^T\bsigmasi (I+\bw\bw^T(L\bsigmas-\bw\bw^T)^{-1})\mathbb{E}vv^T
\big]-Tr\big[
\bsigmasi\mathbb{E}vv^T
\big]\\
=&Tr\big[
\bus\begin{bmatrix}
[I+(\blambdaq-\sigma^2I)((L-1)\blambdaq+\sigma^2I)^{-1}]^2 & 0 \\
0 & I
\end{bmatrix}\blambdas^{-1}\blambdaj\bus^T
\big]-Tr\big[
\blambdas^{-1}\blambdaj
\big]\\
=&Tr\big[
\begin{bmatrix}
[I+(\blambdaq-\sigma^2I)((L-1)\blambdaq+\sigma^2I)^{-1}]^2 & 0 \\
0 & I
\end{bmatrix}\blambdas^{-1}\blambdaj
\big]-Tr\big[
\blambdas^{-1}\blambdaj
\big]\\
=&\sum_{i=1}^q\Big[(1+\frac{\lambda_i-\sigma^2}{(L-1)\lambda_i+\sigma^2})^2-1\Big]\frac{\lambda_i^{(j)}}{\lambda_i},\quad j=1,2,3.
\end{aligned}
\end{equation*}

The excess risk caused by changed of distribution is
\begin{equation*}
\begin{aligned}
&2(\mathcal{L}(\boldsymbol{\Theta_*},\mathcal{D}_j')-\mathcal{L}(\boldsymbol{\Theta_*},\mathcal{D}))\\
=& \log|\bsigmaj|-\log|\bsigmas|+\mathbb{E}_{x'} (x'-\bmus)^T\bsigmasi (x'-\bmus)-\mathbb{E}_{x} (x-\bmus)^T\bsigmasi (x-\bmus)\\
=& \log|\bsigmaj|-\log|\bsigmas|+Tr(\bsigmasi\mathbb{E}_{x'}  (x'-\bmus)(x'-\bmus)^T)-Tr(\bsigmasi \mathbb{E}_{x}(x-\bmus)(x-\bmus)^T)\\
=&\log\Big[\frac{\prod_{i=1}^{d}\lambda_i^{(j)}}{\prod_{i=1}^{d}\lambda_i}\Big]+Tr(\blambdas^{-1}\blambdaj)-Tr(\blambdas^{-1}\blambdas)\\
=&\log\Big[\frac{\prod_{i=1}^{d}\lambda_i^{(j)}}{\prod_{i=1}^{d}\lambda_i}\Big]+\big(\sum_{i=1}^d\frac{\lambda_i^{(j)}}{\lambda_i}-d\big).
\end{aligned}
\end{equation*}
If we further assume that the data lie in a $q$ dimension manifold, by Lemma \ref{lemmaA1}, we have $\sigma^2=0$. $\lambda_i^{(j)}=\lambda_{i}$ and $\mathcal{D}_j'=\mathcal{D}$. Hence the excess risk caused by changed of distribution
\[
\mathcal{L}(\boldsymbol{\Theta_*},\mathcal{D}_j')-\mathcal{L}(\boldsymbol{\Theta_*},\mathcal{D})=0.
\]
The excess risk caused by perturbation is
\begin{equation*}
\begin{aligned}
&2(\mathcal{L}_{gat}(\boldsymbol{\Theta_*},\mathcal{D}_j')-\mathcal{L}(\boldsymbol{\Theta_*},\mathcal{D}_j'))\\
=&\sum_{i=1}^q\Big[(1+\frac{\lambda_i-\sigma^2}{(L-1)\lambda_i+\sigma^2})^2-1\Big]\frac{\lambda_i^{(j)}}{\lambda_i}\\
=&\sum_{i=1}^q\Big[(1+\frac{1}{(L-1)})^2-1\Big]\\
=&\Theta(qL^{-2}).
\end{aligned}
\end{equation*}\qed\\

\subsection{Proof of Thm. \ref{thm:er3}}
Proof: When $q=d$, the inner maximization problem is 
\[
\max_{\|\Delta x\|\leq\varepsilon}\ell(x+\Delta x,\boldsymbol{\mu},\boldsymbol{\Sigma}).
\]
The Lagrangian function is
\begin{equation*}
\begin{aligned}
&\ell(x+\Delta x,\boldsymbol{\mu},\boldsymbol{\Sigma})-\frac{L}{2}(\|\Delta x\|^2-\varepsilon^2)\\
= &\frac{d}{2}\log(2\pi)+\frac{1}{2}\log|\bsigma|+\frac{1}{2}(x-\bmu+\Delta x)^T\bsigma^{-1}(x-\bmu+\Delta x)-\frac{L}{2}(\|\Delta x\|^2-\varepsilon^2).
\end{aligned}
\end{equation*}
Notice that this quadratic objective function is concave when $L$ is larger than the largest eigenvalue of $\bsigma^{-1}$. Calculate the partial derivative with respect to $\Delta x$ and set it to be zero, we have
\begin{equation*}
\begin{aligned}
&\bsigma^{-1}(x-\bmu+\Delta x^*)-L\Delta x^*=0\\
\Leftrightarrow & \Delta x^*= (L\bsigma-I)^{-1}(x-\bmu).
\end{aligned}
\end{equation*}

The excess risk is
\begin{equation*}
\begin{aligned}
&2(\mathcal{L}_{r}(\boldsymbol{\Theta_*},\mathcal{D})-\mathcal{L}(\boldsymbol{\Theta_*},\mathcal{D}))\\
= &\mathbb{E}(v+(L\bsigmas-I)^{-1}v)^T\bsigmasi (v+(L\bsigmas-I)^{-1}v)-\mathbb{E} v^T\bsigmasi v\\
=&Tr\big[
(I+(L\bsigmas-I)^{-1})^T\bsigmasi (I+(L\bsigmas-I)^{-1})\mathbb{E}vv^T
\big]-Tr\big[
\bsigmasi\mathbb{E}vv^T
\big]\\
=&\sum_{i=1}^{d}[(1+\frac{1}{L\lambda_{i}-1})^2-1].
\end{aligned}
\end{equation*}

On the one hand,
\begin{equation}
\label{lower}
\begin{aligned}
&\sum_{i=1}^{d}[(1+\frac{1}{L\lambda_{i}-1})^2-1]\\
\geq &[(1+\frac{1}{L\lambda_{min}-1})^2-1]\\
\geq&\Omega((L\lambda_{min})^{-2}).
\end{aligned}
\end{equation}

On the other hand,
\begin{equation}
\label{upper}
\begin{aligned}
&\sum_{i=1}^{d}[(1+\frac{1}{L\lambda_{i}-1})^2-1]\\
\leq &d[(1+\frac{1}{L\lambda_{min}-1})^2-1]\\
\leq&\mathcal{O}(d(L\lambda_{min})^{-2}).
\end{aligned}
\end{equation}
When $q\leq d$,
$
\max_{\|\Delta x\|\leq\varepsilon}\ell(x+\buq\Delta x,\boldsymbol{\mu},\boldsymbol{\Sigma})=\max_{\|\Delta z\|\leq\varepsilon}\ell(z+\Delta z,\boldsymbol{\mu},\boldsymbol{\Sigma}_q),$ Thm. \ref{thm:er3} is obtain by replacing $\lambda_{min}$ with $\lambda_q$ in Eq. (\ref{lower}) and (\ref{upper}).

\subsection{Proof of Thm. \ref{thm:alp2} and Thm. \ref{thm:robust}}
We first state the general form of Thm. \ref{thm:alp2}.
\begin{theorem}[Optimal solution of generative adversarial training]
	The optimal solution of the modified problem in Eq. (\ref{equ:modify}) is
	\[\boldsymbol{\mu}_{gat}=\boldsymbol{\mu_*}\quad \text{and} \quad
	\boldsymbol{\Sigma}_{gat}=\boldsymbol{U_*}\boldsymbol{\Lambda}^{gat}\boldsymbol{U_*}^T,
	\]
	where 
	\begin{eqnarray*}
	&\lambda_i^{gat}=\frac{1}{4}\Big[2\lambda_i^{(j)}+\frac{4(\lambda_i-\sigma^2)}{L}+2\lambda_i^{(j)}\sqrt{1+\frac{4(\lambda_i-\sigma^2)}{\lambda_i^{(j)}L}}\Big]\ \text{for}\ i=1\leq q\ \text{and}\\
	&\lambda_i^{gat}=\lambda_i^{(j)}\ \text{for}\ i>q.
	\end{eqnarray*}
	$j=1,2,3$ corresponding to strategies 1,2 and 3.
\end{theorem}

By Lemma \ref{lemma:optperb}, the optimal perturbation $\Delta z^*$ is a matrix $M$ times $x-\bmu$. Consider the problem
\begin{equation}
\label{equ:5}
\min_{\boldsymbol{\mu},\boldsymbol{\Lambda}}\max_{\mathbb{E}_{x'}\|M(x'-\bmu)\|=\varepsilon}\mathbb{E}_{x'\sim\mathcal{D}_j'}\ell(x'+\boldsymbol{W}M(x'-\bmu),\boldsymbol{\mu},\boldsymbol{\Sigma}),\quad j=1,2,3.
\end{equation}

\begin{lemma}[optimal perturbation of $M$]
\label{lemma:M}
	Given $\boldsymbol{\Theta}=(\boldsymbol{\mu},\boldsymbol{\Sigma})$ the optimal solution of the inner max problem of \ref{equ:5} is
	\[
 M^*=\boldsymbol{W}^T(L\boldsymbol{\Sigma}-\boldsymbol{W}\boldsymbol{W}^T)^{-1}.
	\]
\end{lemma}
Proof: Consider the problem
\[
\max_{\mathbb{E}\|M(x'-\bmu)\|=\varepsilon}\mathbb{E}\ell(x'+\bw M(x'-\bmu),\bmu,\bsigma)).
\]
The lagrangian function is
\begin{equation*}
\begin{aligned}
\mathbb{E}\big[\ell(x'+\boldsymbol{W}M(x'-\bmu),\boldsymbol{\mu},\boldsymbol{\Sigma})-\frac{L}{2}(\|M(x'-\bmu)\|^2-\varepsilon^2)\big].
\end{aligned}
\end{equation*}
Let $x'-\bmu=v$, Take the gradient with respect to $M$ and set it to be zero, we have
\begin{equation*}
\begin{aligned}
&\frac{\partial}{\partial M}\mathbb{E}\big[\ell(x'+\boldsymbol{W}M(x'-\bmu),\boldsymbol{\mu},\boldsymbol{\Sigma})-\frac{L}{2}(\|M(x'-\bmu)\|^2-\varepsilon^2)\big]
\\
=&\nabla_M \mathbb{E}\big[v^TM\bw^T\bsigma^{-1}v+\frac{1}{2}v^TM\bw^T\bsigma^{-1}\bw M v-Lv^TMMv/2\big]\\
=&\big[\bw^T\bsigma^{-1}+\bw^T\bsigma^{-1}\bw M-LM\big]\mathbb{E}[vv^T]\\
=&0.
\end{aligned}
\end{equation*}
Then we have
\begin{equation*}
\begin{aligned}
M^*=&(L-\bw^T\bsigma^{-1}\bw)^{-1}\bw^T\bsigma^{-1}\\
=&\boldsymbol{W}^T(L\boldsymbol{\Sigma}-\boldsymbol{W}\boldsymbol{W}^T)^{-1}.
\end{aligned}
\end{equation*}
The last equality is due to the Woodbury matrix inversion Lemma. \qed\\

To solve the problem \ref{equ:5}, we need to introduce Danskin's Theorem.
\begin{theorem}[Danskin's Theorem]
Suppose $\phi(x,z): \mathcal{X}\times\mathcal{Z}\rightarrow\mathbb{R}$ is a continuous function of two arguments, where $\mathcal{Z}\subset\mathbb{R}^m$ is compact. Define $f(x) = \max _{z\in\mathcal{Z}}\phi(x, z)$. Then, if for every $z\in\mathcal{Z}$, $\phi(x, z)$ is convex and
differentiable in $x$, and $\partial\phi/\partial x$ is continuous:\\
The subdifferential of $f(x)$ is given by
\[
\partial f(x)=\text{conv}\big\{
\frac{\partial\phi(x,z)}{\partial x},z\in\mathcal{Z}_0(x)
\big\},
\]
where $\text{conv}(\cdot)$ is the convex hull, and $\mathcal{Z}_0(x)$ is
\[
\mathcal{Z}_0(x)=\{
\bar{z}: \phi(x,\bar{z}) = \max \phi(x,z)
\}.
\]
\end{theorem}
If the outer minimization problem is convex and differentiable, we can use any maximizer for the inner maximization problem to find the saddle point. Now we move to the proof of Theorem \ref{thm:alp2}. By Lemma \ref{lemma:M}, we have
\begin{equation*}
\begin{aligned}
M^*=&\boldsymbol{W}^T(L\boldsymbol{\Sigma}-\boldsymbol{W}\boldsymbol{W}^T)^{-1}\\
=& \begin{bmatrix}
\blambdaq-\sigma^2 & 0\\
0&0
\end{bmatrix}^{1/2}\Big(L\blambda- \begin{bmatrix}
\blambdaq-\sigma^2 & 0\\
0&0
\end{bmatrix}\Big)^{-1}\bus^{T}.
\end{aligned}
\end{equation*}
Which is a diagonal matrix $\blambdam$ times $\bus^T$. Let $T=\blambda^{-1}$, $m=\blambda^{-1}\bus^T\bmu$ and $x''=\bus^Tx'$. The optimization problem becomes
\begin{equation}
\begin{aligned}
\label{equ:6}
\min_{m,T}\max_{\blambdam}\quad&\mathbb{E}_{x'\sim\mathcal{D}_j'}\ell(x'+\boldsymbol{W}\blambdam\bus^T(x'-\bmu),m,T)\\
\text{s.t.}\quad&\mathbb{E}_{x'}\|\blambdam\bus(x'-\bmu)\|^2=\varepsilon^2.
\end{aligned}
\end{equation}
Obviously, the inner constraint is compact (by Heine-Borel theorem), we only need to prove the convexity of the outer problem to use Danskin's Theorem. For any $x'$ and $\blambdam$,
\begin{equation*}
\begin{aligned}
&\ell(x'+\boldsymbol{W}\blambdam^T\bus(x'-\bmu),m,T)\\
=&\frac{d}{2}\log(2\pi)+\frac{1}{2}\log|\bsigma|+\frac{1}{2}(x'-\bmu+\bw M(x'-\bmu)^T\bsigma^{-1}(x'-\bmu+\bw M(x'-\bmu).
\end{aligned} 
\end{equation*}
Let $u=\bus^T(x'-\bmu)$, and $A=(I+\begin{bmatrix}
\blambdaq-\sigma^2 & 0\\
0&0
\end{bmatrix}\blambdam)^2$, consider the third term, we have
\begin{equation*}
\begin{aligned}
&\frac{1}{2}\log|\bsigma|+\frac{1}{2}(x'-\bmu+\bw M(x'-\bmu)^T\bsigma^{-1}(x'-\bmu+\bw M(x'-\bmu)\\
=&\frac{1}{2}u^TA^2 T u.
\end{aligned}
\end{equation*}
By \citep{daskalakis2018efficient}, The hessian matrix is
\[
H=\text{Cov}_{z\sim\mathcal{N}(T^{-1}m,(AT)^{-1})}\Big[
\begin{pmatrix}
	\text{vec}(-\frac{1}{2}Azz^T)\\
	z\end{pmatrix},\begin{pmatrix}
	\text{vec}(-\frac{1}{2}Azz^T)\\
	z\end{pmatrix}
\Big]\succeq 0.
\]
Therefore, the outer problem of Eq. (\ref{equ:6}) is a convex problem. By Lemma \ref{lemma:M}, a maximizer of the inner problem is
\begin{equation*}
\begin{aligned}
\blambdam^*=& \begin{bmatrix}
\blambdaq-\sigma^2 & 0\\
0&0
\end{bmatrix}^{1/2}\Big(L\blambda- \begin{bmatrix}
\blambdaq-\sigma^2 & 0\\
0&0
\end{bmatrix}\Big)^{-1}
\end{aligned}.
\end{equation*}
Then 
\[
A=\Big[
I+\begin{bmatrix}
\blambdaq-\sigma^2 & 0\\
0&0
\end{bmatrix}\Big(L\blambda- \begin{bmatrix}
\blambdaq-\sigma^2 & 0\\
0&0
\end{bmatrix}\Big)^{-1}
\Big]^2.
\]
Then the first order derivative (by \citep{daskalakis2018efficient}) is
\[
\nabla_{[T,m]^T}\ell=\begin{bmatrix}
\frac{1}{2}A\blambdaj-\frac{1}{2}T^{-1}\\
AT^{-1}m-A\bus^T\bmus
\end{bmatrix}=0.
\]
From the second equation, we directly have $\bmu_{gat}=\bmus$. From the first equation, for $i> q$, we have
\[
(1+0)^2\lambda_i^{(j)}=\lambda_i^{gat}.
\]
For $i\leq q$, we have
\[
(1+(\lambda_i-\sigma^2)/(L\lambda_i^{gat}-\lambda_i+\sigma^2))^2\lambda_i^{(j)}=\lambda_i^{gat}.
\]
It equivalents to a second order equation of $\sqrt{\lambda_i^{gat}}$
\[
\sqrt{\lambda_i^{gat}}^2-\sqrt{\lambda_i^{(j)}}\sqrt{\lambda_i^{gat}}-\frac{\lambda_i-\sigma^2}{L}=0.
\]
Solving this equation, we obtained
\begin{eqnarray*}
&\lambda_i^{gat}=\frac{1}{4}\Big[2\lambda_i^{(j)}+\frac{4(\lambda_i-\sigma^2)}{L}+2\lambda_i^{(j)}\sqrt{1+\frac{4(\lambda_i-\sigma^2)}{\lambda_i^{(j)}L}}\Big]\ \text{for}\ i=1\leq q\ \text{and}\\
&\lambda_i^{gat}=\lambda_i^{(j)},\ \text{for}\ i>q.
\end{eqnarray*}
Thm. \ref{thm:robust} is obtained by replacing $W$ with $\boldsymbol{U}_q$.\qed

\clearpage
\section{Addtional Experiments}
\subsection{Training Settings}
\label{B1}
On MNIST, we use LeNet5 for the classifier and 2 layers MLP (with hidden size 256 and 784) for the encoder and decoder of conditional VAE. For standard training of the classifier, we use 30 epochs, batch size 128, learning rate $10^{-3}$, and weight decay $5\times10^{-4}$. For the CVAE, we use 20 epochs, learning rate $10^{-3}$, batch size 64, and latent size 10. For standard adversarial training and, we use $\varepsilon=0.3$ for FGSM and PGD. in PGD, we use 40 steps for the inner part. Adversarial training start after 10 epochs standard training. For generative adversarial training, we use $\varepsilon=1$ in the latent space with FGSM and PGD. We use 40 steps PGD for latent space adversarial training. Adversarial training start after 10 epoches standard training. In the attack part, we use the same $\varepsilon$ as the adversarial training part.

On CIFAR-10, we use ResNet32 for the classifier and 4 layers CNN for the encoder and decoder of conditional VAE. For standard training of the classifier, we use 200 epochs, batch size 128. The learning rate schedule is 0.1 for the first 100 epochs, 0.01 for the following 50 epochs, and 0.001 for the last 50 epochs. The weight decay is $5\times10^{-4}$. For the CVAE, we use 200 epochs, learning rate $10^{-3}$, batch size 64, and latent size 128.

For standard adversarial training, we use $\varepsilon=8/255$ for FGSM and PGD. in PGD, we use 10 steps for the inner part. For generative adversarial training, we use $\varepsilon=0.1$ in the latent space with FGSM and PGD. Since we see that the modeling power of VAE in CIFAR10 is not good enough. For each of the image, the encode variance is very small. When we add a small perturbation to the encode mean value, the output image are blured. Hence we only use a small $\varepsilon=0.1$. In the attack part, we use $\varepsilon=8/255$ for norm-based attacks and $\varepsilon=0.1$ for generative attack on the test set.

\paragraph{CIFAR-100} The training hyperparameters are the same as that on CIFAR-10. In Table \ref{t100}, we provide the experiments results on CIFAR-100. We ca get the same conclusion as in the MNIST and CIFAR-10. Standard adversarial training cannot defense generative attack, and vice versa. Notice that the modeling power of condition-VAE is not good enough for CIFAR-100. Therefore, we use a more powerful StyleGAN for ImageNet.

\begin{table*}[htbp]
	\centering
	\caption{Test accuracy of different defense algorithms (PGD-AT, GPGD-AT, and joint AT) against different attacks (regular attacks (FGSM, PGD) and generative attack (GFGSM, GPGD)) on CIFAR-100.}
	\resizebox{\linewidth}{!}{%
	\begin{tabular}{cccccc}
		\hline
		CIFAR-100 & clean data & FGSM-Attack & PGD-Attack& VAE-FGSM-Attack& VAE-PGD-Attack\\
		\hline
		Std training &  67.26\%&9.13\%&7.31\%&1.67\%& 0.34\%\\
		PGD-AT & 50.85\%&31.49\%&27.27\%&6.41\%&4.18\%\\
		VAE-PGD-AT&45.21\%&3.45\%&1.21\%&12.45\%&10.75\%\\
		joint-PGD-AT &48.34\%&30.81\%&25.55\%&12.01\%&10.12\%\\
		\hline
	\end{tabular}}
	\label{t100}
\end{table*}

\paragraph{ImageNet} We adopt the setting of dual manifold adversarial training (DMAT) \citep{lin2020dual} in our experiments. DMAT is an algorithm using both on-manifold and off-manifold adversarial examples to train a neural network. The off-manifold adversarial examples are $\ell_\infty$ PGD adversarial examples. The on-manifold adversarial examples are crafted by a StyleGAN \citep{karras2019style} trained on ImageNet. The difference between the algorithms we used and DMAT is the generative model, CVAE, and StyleGAN. StyleGAN is better on ImageNet because of the modeling power. In Table \ref{t2}, we show the results of DMAT on ImageNet. We can see that, on-manifold adversarial training (GPGD-AT) is not able to defend PGD-Attack, with 0.03\% robust test accuracy, and vice versa. Joint-AT can achieve comparable robust test accuracy to the single-trained models. Therefore, the results of DMAT support our analysis. Conversely, our theory gives an analysis of the performance of DMAT.

\subsection{Eigenspace Adversarial Training}
\label{B2}
 \begin{figure*}[htbp]
\centering
\subfigure[]{
\begin{minipage}[t]{0.24\linewidth}
\centering
\includegraphics[width=1.2in]{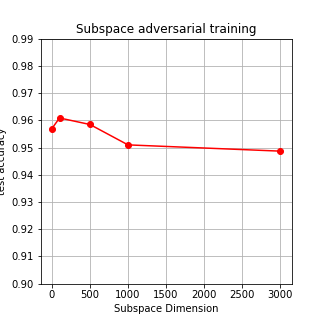}
\end{minipage}%
}%
\subfigure[]{
\begin{minipage}[t]{0.24\linewidth}
\centering
\includegraphics[width=1.2in]{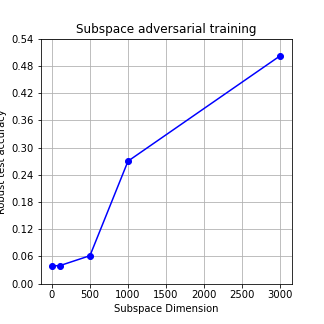}
\end{minipage}%
}%
\subfigure[]{
\begin{minipage}[t]{0.24\linewidth}
\centering
\includegraphics[width=1.2in]{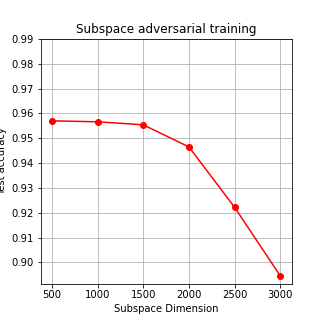}
\end{minipage}
}%
\subfigure[]{
\begin{minipage}[t]{0.24\linewidth}
\centering
\includegraphics[width=1.2in]{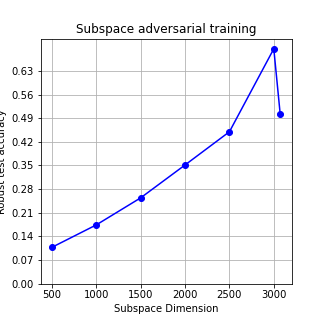}
\end{minipage}
}%
\centering
\caption{Standard and robust test accuracies of subspace adversarial training on CIFAR-10. The x-axis is the dimension of the restricted subspace of adversarial training. The red lines are the standard test accuracies and the blue lines are the robust test accuracies for (full space)  PGD-10 $\ell_2$ attacks. (a) and (b) are the results of on-manifold adversarial training, i.e. adversarial training restricted in the subspace spanned by the top 500, 1000, $\cdots$, 3000 eigenvectors; (c) and (d) are the results of off-manifold adversarial training.}
\label{fig3}
\end{figure*}

In Figure \ref{fig3} (a) and (b), we give the experimental results of on-manifold adversarial training, i.e. adversarial training restricted in the subspace spanned by the top eigenvectors. The red lines are the standard test accuracies and the blue lines are the robust test accuracies against PGD-10 $\ell_2$ attacks. (c) and (d) are the results of off-manifold adversarial training.

\paragraph{Bridging the generalization-robustness trade-off} The range of the x-axis is from 0 to 3072. Notice that if the number of subspace dimension is $0$, no perturbation is allowed. It equivalents to the standard training case. If the number of subspace dimension is $3072$, the perturbation $\Delta x$ can be chosen in the whole norm ball $\{\|\cdot\|\leq\varepsilon\}$. It equivalents to the regular adversarial training case. Therefore, subspace adversarial training is a middle situation between standard training and adversarial training.

\paragraph{On-manifold adversarial training improves generalization} In Figure \ref{fig3} (a), the best test accuracy is 96.17\%, achieved in subspace adversarial training restricted to the top 100 eigenvectors. In Figure \ref{fig3} (c), we can see that off-manifold adversarial training cannot obtain a better test accuracy than standard training. It is because on-manifold adversarial examples will cause a small distributional shift, the negative effect of distributional shift is smaller than the positive effect of data augmentation. But for off-manifold adversarial examples, the negative effect of off-manifold adversarial examples is larger.

\paragraph{Eigenspace adversarial training cannot improves robustness} In Figure \ref{fig3}, we show the experiments of Off-manifold adversarial training. We can see that eigenspace adversarial training cannot improves robustness.

 \begin{figure*}[htbp]
\centering
\subfigure[]{
\begin{minipage}[t]{0.24\linewidth}
\centering
\includegraphics[width=1.2in]{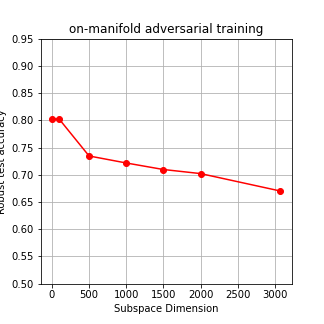}
\end{minipage}%
}%
\subfigure[]{
\begin{minipage}[t]{0.24\linewidth}
\centering
\includegraphics[width=1.2in]{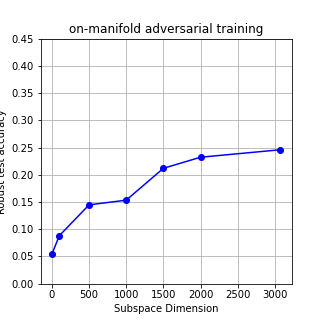}
\end{minipage}%
}%
\subfigure[]{
\begin{minipage}[t]{0.24\linewidth}
\centering
\includegraphics[width=1.2in]{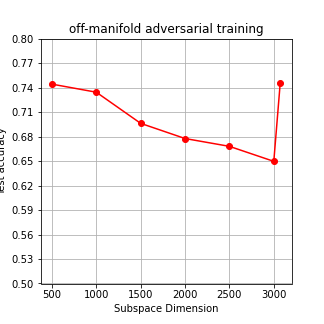}
\end{minipage}
}%
\subfigure[]{
\begin{minipage}[t]{0.24\linewidth}
\centering
\includegraphics[width=1.2in]{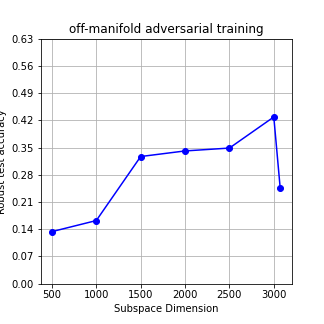}
\end{minipage}
}%
\centering
\caption{Standard and robust test accuracies of subspace adversarial training on CIFAR-100. The red lines are the standard test accuracies and the blue lines are the robust test accuracies for (full space)  PGD-10 $\ell_2$ attacks. (a) and (b) are the results of on-manifold adversarial training; (c) and (d) are the results of off-manifold adversarial training.}
\label{fig5}
\end{figure*}
\paragraph{Eigenspace adversarial training on CIFAR-100} As it is shown in Figure \ref{fig5}, on-manifold adversarial training (top 100 dimensions) get the same test accuracy as standard training. Off-manifold training (bottom 3000 dimensions) get larger robust accuracy (42.89\%) than standard adversarial training (24.60\%). The results are similar to those in CIFAR-10.

\subsection{Ablation study of adversarial distribution shift}
\label{sec:B3}
\begin{figure}[t]
	\centering
	\scalebox{0.9}{
		\begin{minipage}[htp]{0.01\linewidth}
			\centering
			\rotatebox{90}{\text{Class 0}}
		\end{minipage}
		\subfigure[]{
			\begin{minipage}[htp]{0.23\linewidth}
				\centering
				\includegraphics[width=1.3in]{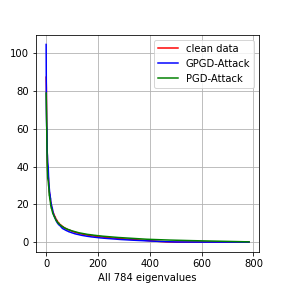}
			\end{minipage}%
		}
		\subfigure[]{
			\begin{minipage}[htp]{0.23\linewidth}
				\centering
				\includegraphics[width=1.3in]{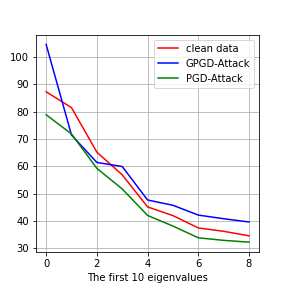}
			\end{minipage}
		}
		\subfigure[]{
			\begin{minipage}[htp]{0.23\linewidth}
				\centering
				\includegraphics[width=1.3in]{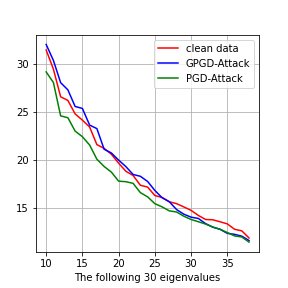}
			\end{minipage}
		}
		\subfigure[]{
			\begin{minipage}[htp]{0.23\linewidth}
				\centering
				\includegraphics[width=1.3in]{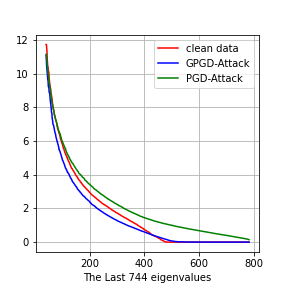}
			\end{minipage}
	}}
	\scalebox{0.9}{
		
		\begin{minipage}[htp]{0.01\linewidth}
			\centering
			\rotatebox{90}{\text{Class 1}}
		\end{minipage}%
		\subfigure[]{
			\begin{minipage}[htp]{0.23\linewidth}
				\centering
				\includegraphics[width=1.3in]{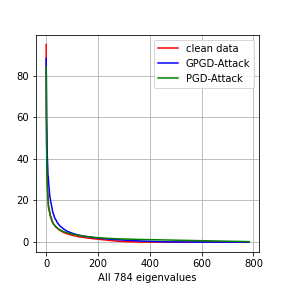}
			\end{minipage}%
		}
		\subfigure[]{
			\begin{minipage}[htp]{0.23\linewidth}
				\centering
				\includegraphics[width=1.3in]{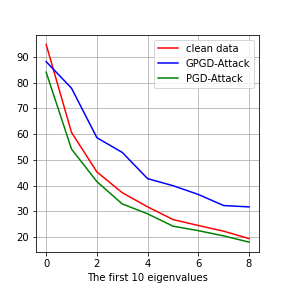}
			\end{minipage}
		}
		\subfigure[]{
			\begin{minipage}[htp]{0.23\linewidth}
				\centering
				\includegraphics[width=1.3in]{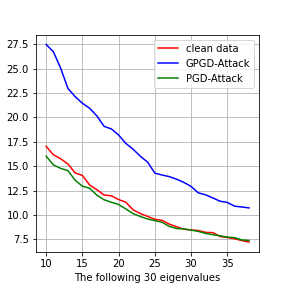}
			\end{minipage}
		}
		\subfigure[]{
			\begin{minipage}[htp]{0.23\linewidth}
				\centering
				\includegraphics[width=1.3in]{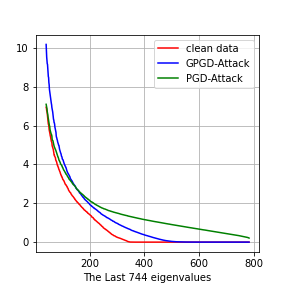}
			\end{minipage}
	}}
	\scalebox{0.9}{
	
	\begin{minipage}[htp]{0.01\linewidth}
		\centering
		\rotatebox{90}{\text{Class 2}}
	\end{minipage}%
	\subfigure[]{
		\begin{minipage}[htp]{0.23\linewidth}
			\centering
			\includegraphics[width=1.3in]{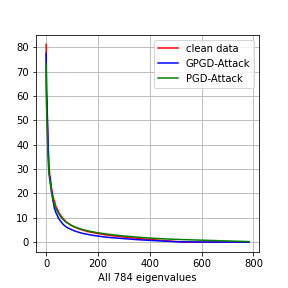}
		\end{minipage}%
	}
	\subfigure[]{
		\begin{minipage}[htp]{0.23\linewidth}
			\centering
			\includegraphics[width=1.3in]{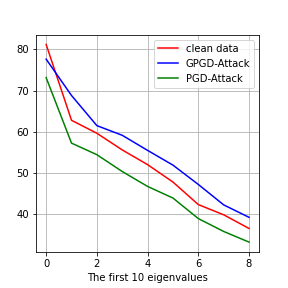}
		\end{minipage}
	}
	\subfigure[]{
		\begin{minipage}[htp]{0.23\linewidth}
			\centering
			\includegraphics[width=1.3in]{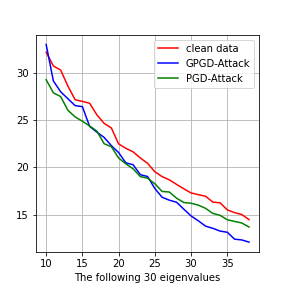}
		\end{minipage}
	}
	\subfigure[]{
		\begin{minipage}[htp]{0.23\linewidth}
			\centering
			\includegraphics[width=1.3in]{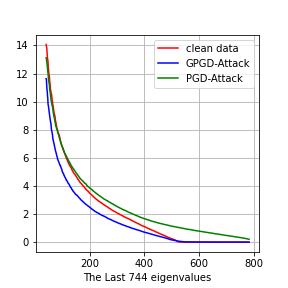}
		\end{minipage}
}}
	\scalebox{0.9}{
	
	\begin{minipage}[htp]{0.01\linewidth}
		\centering
		\rotatebox{90}{\text{Class 3}}
	\end{minipage}%
	\subfigure[]{
		\begin{minipage}[htp]{0.23\linewidth}
			\centering
			\includegraphics[width=1.3in]{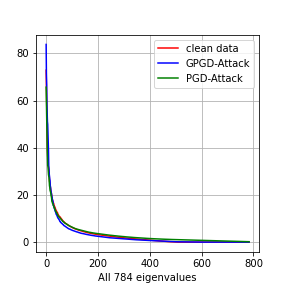}
		\end{minipage}%
	}
	\subfigure[]{
		\begin{minipage}[htp]{0.23\linewidth}
			\centering
			\includegraphics[width=1.3in]{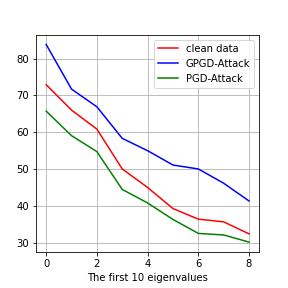}
		\end{minipage}
	}
	\subfigure[]{
		\begin{minipage}[htp]{0.23\linewidth}
			\centering
			\includegraphics[width=1.3in]{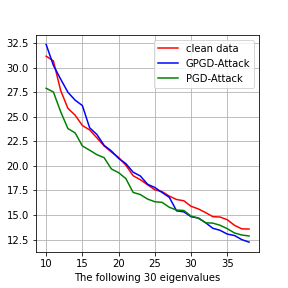}
		\end{minipage}
	}
	\subfigure[]{
		\begin{minipage}[htp]{0.23\linewidth}
			\centering
			\includegraphics[width=1.3in]{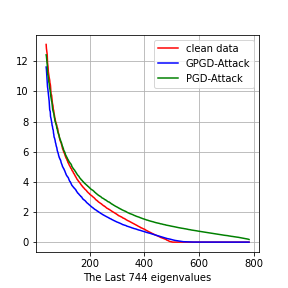}
		\end{minipage}
}}
	\scalebox{0.9}{
	
	\begin{minipage}[htp]{0.01\linewidth}
		\centering
		\rotatebox{90}{\text{Class 4}}
	\end{minipage}%
	\subfigure[]{
		\begin{minipage}[htp]{0.23\linewidth}
			\centering
			\includegraphics[width=1.3in]{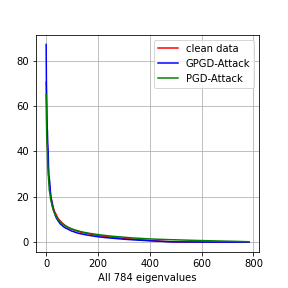}
		\end{minipage}%
	}
	\subfigure[]{
		\begin{minipage}[htp]{0.23\linewidth}
			\centering
			\includegraphics[width=1.3in]{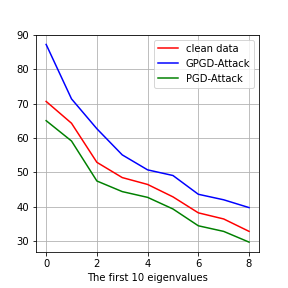}
		\end{minipage}
	}
	\subfigure[]{
		\begin{minipage}[htp]{0.23\linewidth}
			\centering
			\includegraphics[width=1.3in]{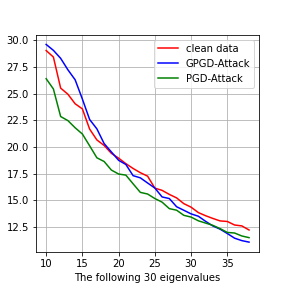}
		\end{minipage}
	}
	\subfigure[]{
		\begin{minipage}[htp]{0.23\linewidth}
			\centering
			\includegraphics[width=1.3in]{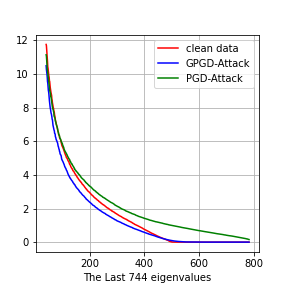}
		\end{minipage}
}}
	\centering
	\vskip -0.15in
	\caption{Adversarial distributional shift on MNIST for all the 10 classes.}
	\label{alleigen}
	\vskip -0.1in
\end{figure}

\begin{figure}[t]
	\centering
	\scalebox{0.9}{
		\begin{minipage}[htp]{0.01\linewidth}
			\centering
			\rotatebox{90}{\text{Class 5}}
		\end{minipage}
		\subfigure[]{
			\begin{minipage}[htp]{0.23\linewidth}
				\centering
				\includegraphics[width=1.3in]{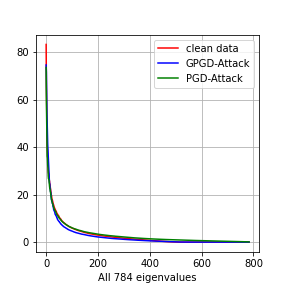}
			\end{minipage}%
		}
		\subfigure[]{
			\begin{minipage}[htp]{0.23\linewidth}
				\centering
				\includegraphics[width=1.3in]{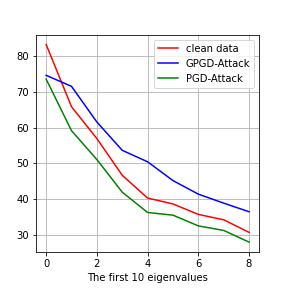}
			\end{minipage}
		}
		\subfigure[]{
			\begin{minipage}[htp]{0.23\linewidth}
				\centering
				\includegraphics[width=1.3in]{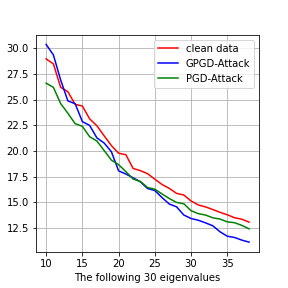}
			\end{minipage}
		}
		\subfigure[]{
			\begin{minipage}[htp]{0.23\linewidth}
				\centering
				\includegraphics[width=1.3in]{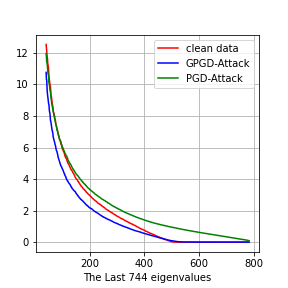}
			\end{minipage}
	}}
	\scalebox{0.9}{
		
		\begin{minipage}[htp]{0.01\linewidth}
			\centering
			\rotatebox{90}{\text{Class 6}}
		\end{minipage}%
		\subfigure[]{
			\begin{minipage}[htp]{0.23\linewidth}
				\centering
				\includegraphics[width=1.3in]{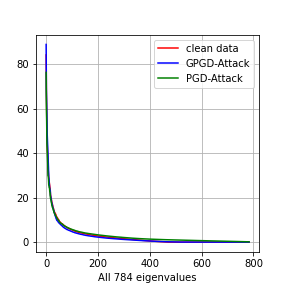}
			\end{minipage}%
		}
		\subfigure[]{
			\begin{minipage}[htp]{0.23\linewidth}
				\centering
				\includegraphics[width=1.3in]{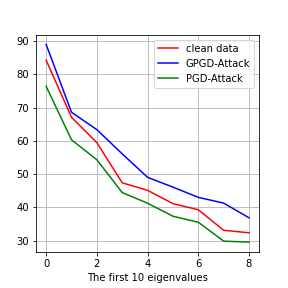}
			\end{minipage}
		}
		\subfigure[]{
			\begin{minipage}[htp]{0.23\linewidth}
				\centering
				\includegraphics[width=1.3in]{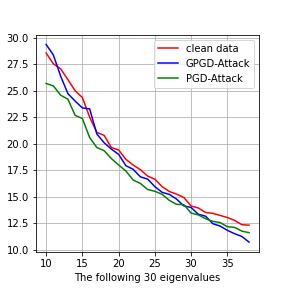}
			\end{minipage}
		}
		\subfigure[]{
			\begin{minipage}[htp]{0.23\linewidth}
				\centering
				\includegraphics[width=1.3in]{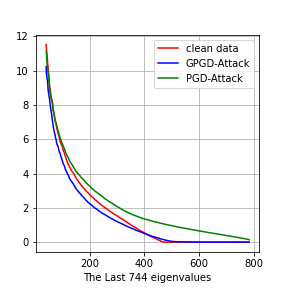}
			\end{minipage}
	}}
	\scalebox{0.9}{
		
		\begin{minipage}[htp]{0.01\linewidth}
			\centering
			\rotatebox{90}{\text{Class 7}}
		\end{minipage}%
		\subfigure[]{
			\begin{minipage}[htp]{0.23\linewidth}
				\centering
				\includegraphics[width=1.3in]{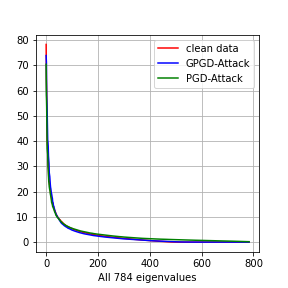}
			\end{minipage}%
		}
		\subfigure[]{
			\begin{minipage}[htp]{0.23\linewidth}
				\centering
				\includegraphics[width=1.3in]{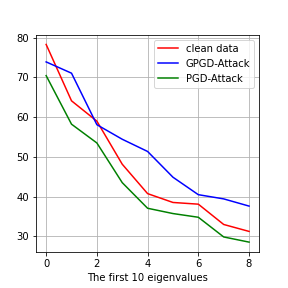}
			\end{minipage}
		}
		\subfigure[]{
			\begin{minipage}[htp]{0.23\linewidth}
				\centering
				\includegraphics[width=1.3in]{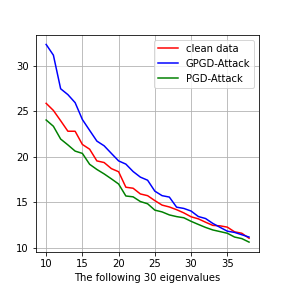}
			\end{minipage}
		}
		\subfigure[]{
			\begin{minipage}[htp]{0.23\linewidth}
				\centering
				\includegraphics[width=1.3in]{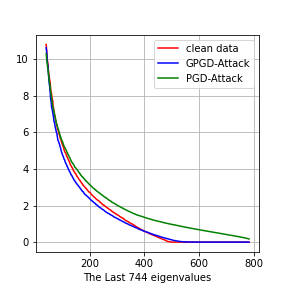}
			\end{minipage}
	}}
	\scalebox{0.9}{
		
		\begin{minipage}[htp]{0.01\linewidth}
			\centering
			\rotatebox{90}{\text{Class 8}}
		\end{minipage}%
		\subfigure[]{
			\begin{minipage}[htp]{0.23\linewidth}
				\centering
				\includegraphics[width=1.3in]{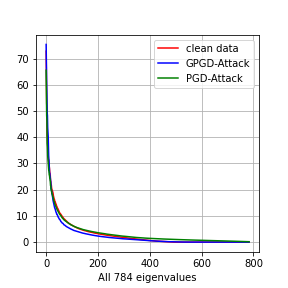}
			\end{minipage}%
		}
		\subfigure[]{
			\begin{minipage}[htp]{0.23\linewidth}
				\centering
				\includegraphics[width=1.3in]{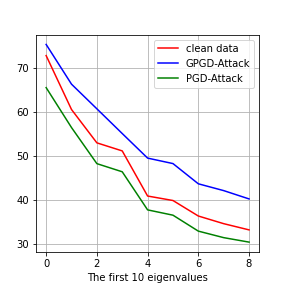}
			\end{minipage}
		}
		\subfigure[]{
			\begin{minipage}[htp]{0.23\linewidth}
				\centering
				\includegraphics[width=1.3in]{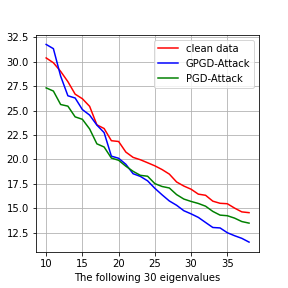}
			\end{minipage}
		}
		\subfigure[]{
			\begin{minipage}[htp]{0.23\linewidth}
				\centering
				\includegraphics[width=1.3in]{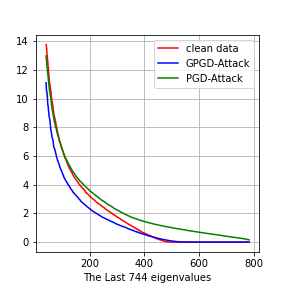}
			\end{minipage}
	}}
	\scalebox{0.9}{
		
		\begin{minipage}[htp]{0.01\linewidth}
			\centering
			\rotatebox{90}{\text{Class 9}}
		\end{minipage}%
		\subfigure[]{
			\begin{minipage}[htp]{0.23\linewidth}
				\centering
				\includegraphics[width=1.3in]{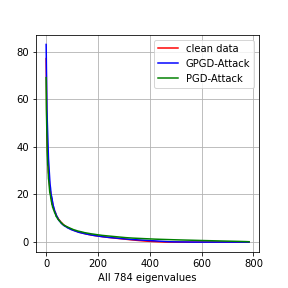}
			\end{minipage}%
		}
		\subfigure[]{
			\begin{minipage}[htp]{0.23\linewidth}
				\centering
				\includegraphics[width=1.3in]{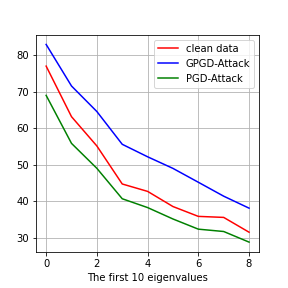}
			\end{minipage}
		}
		\subfigure[]{
			\begin{minipage}[htp]{0.23\linewidth}
				\centering
				\includegraphics[width=1.3in]{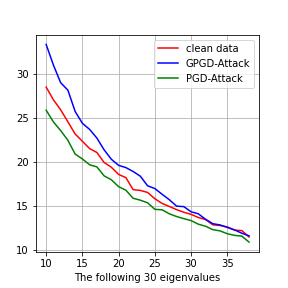}
			\end{minipage}
		}
		\subfigure[]{
			\begin{minipage}[htp]{0.23\linewidth}
				\centering
				\includegraphics[width=1.3in]{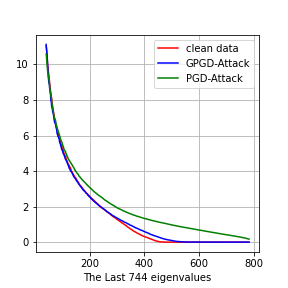}
			\end{minipage}
	}}
	\centering
	\vskip -0.15in
	\caption{Adversarial distributional shift on MNIST for all the 10 classes.}
	\label{alleigen2}
	\vskip -0.1in
\end{figure}

We plot the adversarial distribution shift of all the classes in this section, see Figure \ref{alleigen} and \ref{alleigen2}. For all the 10 classes, We can see that adversarial training will amplify the small eigenvalues and generative adversarial training will amplify the large eigenvalues.

\end{document}